\documentclass[1p,review]{elsarticle}

\usepackage{amsmath}
\usepackage{amssymb}

\usepackage{lineno}
\usepackage{hyperref}
\usepackage{algorithm}
\usepackage{algorithmic}
\usepackage[figuresright]{rotating}
\usepackage{subfigure}
\usepackage{multirow}
\usepackage[capitalise, noabbrev]{cleveref}

\usepackage{soul}
\usepackage{xcolor}
\usepackage{caption}
\usepackage{flushend}

\usepackage{booktabs}
\usepackage{array}
\usepackage{makecell}
\usepackage{graphbox}

\usepackage{ulem}
 
\newcommand{\specialcell}[2][c]{%
  \begin{tabular}[#1]{@{}c@{}}#2\end{tabular}}
  
\newcolumntype{P}[1]{>{\centering\arraybackslash}p{#1}}
\newcolumntype{M}[1]{>{\centering\arraybackslash}m{#1}}

\journal{Knowledge-Based Systems}

\begin{document}
\begin{frontmatter}






%\dochead{Cabecera artaculo}

%-------------------------------------------------------------------------
\title{AI-KD: Adversarial learning and Implicit regularization for self-Knowledge Distillation}


\author[KAIST,SEMCO]{Hyungmin Kim}
\author[DFKI,RPTU]{Sungho Suh\corref{mycorrespondingauthor}}
\cortext[mycorrespondingauthor]{Corresponding author}
\ead{sungho.suh@dfki.de}
\author[KAIST]{Sunghyun Baek}
\author[SEMCO]{Daehwan Kim}
\author[SEMCO]{Daun Jeong}
\author[SEMCO]{Hansang Cho}
\author[KAIST]{Junmo Kim}

\address[KAIST]{School of Electrical Engineering, Korea Advanced Institute of Science and Technology (KAIST), Daejeon 34141, South Korea}
\address[SEMCO]{Samsung Electro-Mechanics, Suwon 16674, South Korea}
\address[DFKI]{German Research Center for Artificial Intelligence (DFKI), 67663 Kaiserslautern, Germany}
\address[RPTU]{Department of Computer Science, RPTU Kaiserslautern-Landau, 67663 Kaiserslautern, Germany}
%-------------------------------------------------------------------------
\begin{abstract}
We present a novel adversarial penalized self-knowledge distillation method, named adversarial learning and implicit regularization for self-knowledge distillation (AI-KD), which regularizes the training procedure by adversarial learning and implicit distillations. Our model not only distills the deterministic and progressive knowledge which are from the pre-trained and previous epoch predictive probabilities but also transfers the knowledge of the deterministic predictive distributions using adversarial learning. The motivation is that the self-knowledge distillation methods regularize the predictive probabilities with soft targets, but the exact distributions may be hard to predict. Our proposed method deploys a discriminator to distinguish the distributions between the pre-trained and student models while the student model is trained to fool the discriminator in the trained procedure. Thus, the student model not only can learn the pre-trained model's predictive probabilities but also align the distributions between the pre-trained and student models. We demonstrate the effectiveness of the proposed method with network architectures on multiple datasets and show the proposed method achieves better performance than existing approaches.
\end{abstract}

%-------------------------------------------------------------------------

\end{frontmatter}

\linenumbers
\nolinenumbers

%-------------------------------------------------------------------------
\section{Introduction}
\label{sec:introduction}
Knowledge distillation (KD)~\cite{hinton2015kd} scheme aims to efficient model compression to transfer knowledge from the larger deep teacher network into the lightweight student network.
The method usually utilizes the Kullback-Leibler (KL) divergence between the softened probability distributions of the teacher and the student network with the temperature scaling parameter.
However, KD has an inherent drawback which is a drop in accuracy by transferring compacted information, so it is not always the effectual approach where there are significant gaps in the quantity of parameters between the teacher and the student models and where the models are heterogeneously different.
Also, it is hard to tell which intermediate features of the teacher are the best selection and where to transfer to features of the student. On the other hand, the multiple selected features may not represent better knowledge than a single feature of the layer.
Thus, various approaches apply neural architecture search and meta-learning to handle the selection, which features and intermediate layers of the teacher and the student models, for better knowledge transfer.
In this manner, Jang and Shin~\cite{jang2019learning}, and Mirzadeh and Ghasemzadeh~\cite{mirzadeh2020improved} proposed utilizing meta networks and bridge models to alleviate differences from the network models, respectively.

\begin{figure*}[!t]
    \centering
    \resizebox{1.0\linewidth}{!}{
    \setlength{\tabcolsep}{5pt}
    \begin{tabular}{cc}
        \includegraphics[width=0.7\linewidth]{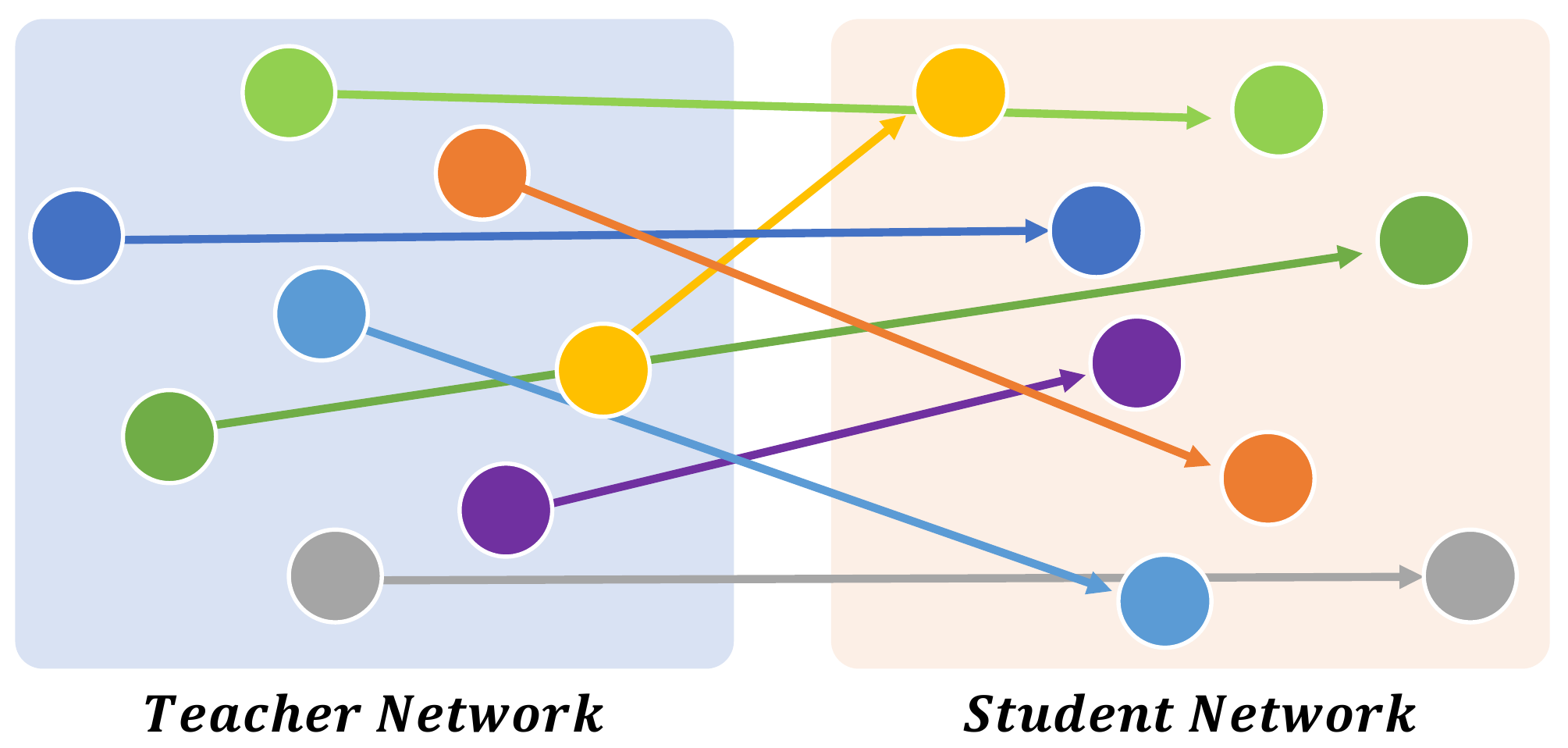} &        \includegraphics[width=0.7\linewidth]{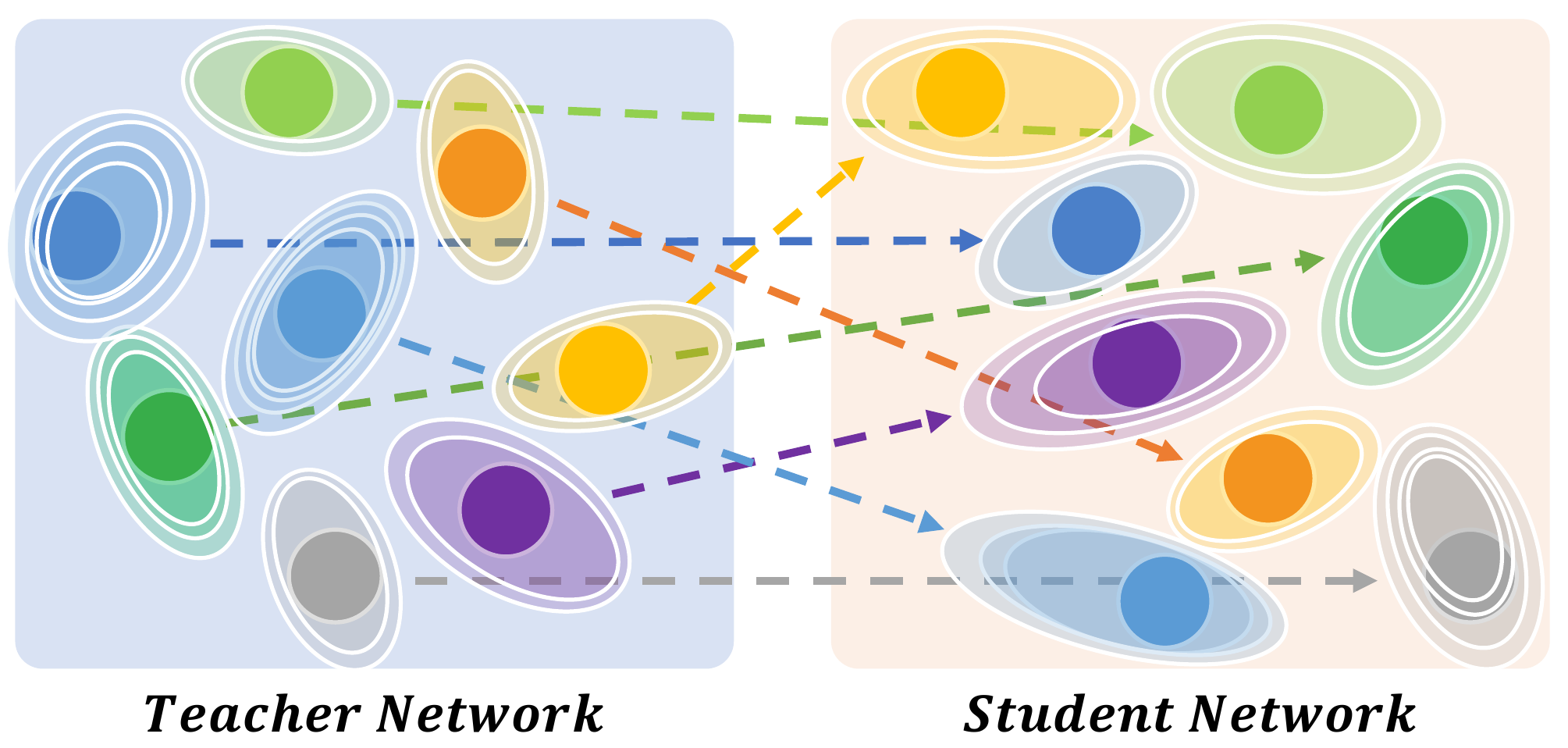}\\
        (a) Conventional Knowledge Distillation &(b) Adversarial Implicit Knowledge Distillation
    \end{tabular}}
    \caption{Concept comparison with conventional knowledge distillation methods. Each point denotes a latent vector represented from the different input data. (a) The solid line indicates that the conventional knowledge distillation is trained to minimize the distance between the latent vector point in the teacher network and the corresponding latent vector point in the student network. (b) The dashed line indicates that AI-KD is trained to minimize the distance between the distribution of the latent vectors in the teacher network and the distribution of the corresponding latent vectors in the student network.}
    \label{fig:concept}
\end{figure*}

On the other hand, self-knowledge distillation (Self-KD) involves training the network itself as a teacher network, aiming to regularize the model for improved generalization and to prevent overfitting using only the student model. In contrast to other KD techniques, Self-KD does not explicitly consider network heterogeneity, as it transfers knowledge within a singular network. Self-KD, serving as a regularization technique, efficiently alleviates overfitting issues and improves generalization performance.
Feature-based Self-KD approaches are effective generalization techniques, utilizing the rich representative capacity of features. However, most of the existing methods minimize the distance between features using metrics like $L_2$ distance or cosine similarity, often requiring additional network modifications. Logit-based Self-KD approaches, on the other hand, offer an intuitive alternative that does not require explicit model adjustments. They work by focusing on the logits, the raw output of the neural networks, simplifying the knowledge transfer process. Despite their initially perceived capacity limitations, recent logit-based approaches have overcome such challenges, demonstrating superior generalization compared to the feature-based approaches. Teacher-free KD (TF-KD)~\cite{yuan2020revisiting} challenged the notion that a powerful and well-trained teacher is an essential component for knowledge transfer. Instead, they revealed that a pre-trained student model could serve as the teacher model, successfully preventing overfitting and presenting comparable performance to conventional KD.
Progressive Self-KD (PS-KD)~\cite{Kim_2021_ICCV} proposed a unique approach by distilling knowledge through the generation of soft labels using historical predictive distributions from the last epoch. However, this method often regularizes the predictive probabilities with soft targets, and predicting exact distributions can be challenging, as depicted in \cref{fig:concept} (a). This focus on mimicking predictive probability information through direct alignment methods limits the potential to reduce distributional differences between models.
%-------------------------------------------------------------------------

%-------------------------------------------------------------------------
In this paper, we propose a novel logit-based Self-KD method via adversarial learning, coined
adversarial learning and implicit regularization for self-knowledge distillation (AI-KD).
The proposed method learns implicitly distilled deterministic and progressive knowledge from the pre-trained model and the previous epoch student model, and aligns the distributions of the student model to the distributions of the pre-trained using adversarial learning, as shown in Figure~\ref{fig:concept} (b).
The pre-trained model is the baseline of the dataset and is trained from scratch following standard training parameters.
To achieve the object, EM distance in WGAN is utilized to allow the discriminator to utilize a scalar value, which improves gradient delivery and enhances learning for both the discriminator and the student model. Due to these benefits,
our proposed method distills knowledge by considering the probability distribution of networks so that the networks align not directly, leading to better model generalization.
To verify the generalization effectiveness, we evaluate the proposed method using ResNet~\cite{he2016deep, he2016identity} and DenseNet~\cite{huang2017densely} on coarse and fine-grained datasets.
In addition, we demonstrate compatibility with recent image augmentation methods to provide broad applicability with other regularization techniques.

The main contributions of the proposed method can be summarized as follows.
%-------------------------------------------------------------------------
\begin{itemize}
    \item We propose a novel adversarial Self-KD, named AI-KD, between the pre-trained and the student models.
    Our method provides the student model can align to the predictive probability distributions of the pre-trained models by adversarial learning.
    \item For training stability, AI-KD distills implicit knowledge of deterministic and progressive from the pre-trained and the previous epoch model, respectively.
    \item In the experiments, AI-KD is evaluated to show the performance of the generalization on various public datasets. We also verify the compatibility of AI-KD with the data augmentations.
\end{itemize}

The rest of the paper is organized as follows. Section~\ref{sec:relatedworks} introduces related work. Section~\ref{sec:method} provides the details of the proposed method. Section~\ref{sec:experimentalresults} presents qualitative and quantitative experimental results with a variety of network architectures on multiple datasets, including coarse datasets and fine-grained datasets. Finally, Section~\ref{sec:conclusion} concludes the paper and addresses future works.

%-------------------------------------------------------------------------
\section{Related Works}
\label{sec:relatedworks}
%-------------------------------------------------------------------------
\subsection{Self-knowledge distillation}
Self-KD is one of the KD approaches that utilize either the same network as the teacher and the student or a homogeneous network with the same size. In this perspective, its primary focus is on improving generalization performance rather than model compression. One of the Self-KD approaches uses multiple auxiliary heads and requires extensive modifications to the network architecture.
These approaches, including Be Your Own Teacher (BYOT)~\cite{Wang_Li_2021} and Harmonized Dense KD (HD-KD)~\cite{Wang_Li_2021}, lead to increased implementation complexity and require complex transformations, such as meta-learning, to effectively convey essential knowledge from the teacher network to the student.
Another approach is utilizing the input-based method. Data-distortion guided Self-KD (DDGSD)~\cite{xu2019data} focuses on constraining consistent outputs across an image distorted differently.
Yun and Shin~\cite{Yun_2020_CVPR} proposed a class-wise Self-KD (CS-KD) that distills the knowledge of a model itself to force the predictive distributions between different samples in the same class by minimizing KL divergence. 
PS-KD~\cite{Kim_2021_ICCV}, Memory-replay KD (Mr-KD)~\cite{wang2021memory}, Self distillation from last mini-batch (DLB)~\cite{xu2022dlb}, and Snapshot~\cite{Yang_2019_CVPR} proposed progressive knowledge distillation methods that distill the knowledge of a model itself as a virtual teacher by utilizing the past predictive distributions at the last epoch, the last minibatch, and the last iteration, respectively.
Consequently, these methods aim to generate appropriate soft labels including discriminative information from previous epochs. However, Zipf$'$s label smoothing (ZipfsLs)~\cite{liang2022efficient}
suggests generating soft labels that follow the Zipf distribution. Ambiguity-aware robust teacher KD~\cite{cho2023ambiguity}
focuses on finding ambiguous samples, which are specified by two different teacher models. These ambiguous samples are refined to soft labels to minimize the distances between the predicted probabilities of the student model and those of the teachers.
Born again neural networks (BANs)~\cite{pmlr-v80-furlanello18a} trains a student network to use as a teacher for the next generations, iteratively. In these manners, it performs multiple generations of ensembled Self-KD. 
In contrast, \cite{yuan2020revisiting} proposed the TF-KD that uses a pre-trained student network as a teacher network for a single generation.
This method aligns directly between the teacher and student model so that the student model only mimics the teacher model.
Moreover, Teacher-free feature distillation (TF-FD)~\cite{li2022self}
proposes to reuse channel-wise and layer-wise discriminative features within the student to provide knowledge similar to that of the teacher without an extra model or complex transformation. And it leads to emphasizing the aligning features of the student model with the teacher. In a different approach from the aforementioned, Sliding Cross Entropy (SCE)~\cite{lee2022sliding}
adopts the modified cross entropy considered inter-class relations.

%-------------------------------------------------------------------------
\subsection{Adversarial learning}
Generative adversarial network (GAN)~\cite{goodfellow2014generative} is a min-max game between a generator and a discriminator to approximate the probability distributions and generate real-like data.
The goal of the generator is to deceive the discriminator into fake determining the results as the real samples. The generator is trained to generate the real data distribution, while the discriminator is trained to distinguish the real samples from the fake samples generated by the generator. 
There are various works proposed to overcome the well-known issue in GAN, which is an unstable training procedure due to the gradient vanishing problem in the training procedure.
Wasserstein GAN (WGAN)~\cite{arjovsky2017wasserstein} and Wasserstein GAN with gradient penalty (WGAN-GP)~\cite{gulrajani2017improved} alleviates the issue using Earth-Mover's (EM) distance and adopts the weight clipping method for training stability.

Recently, adversarial learning has been studied to utilize for KD.
One approach addressed training a generator to synthesize input images by the teacher's features, and then the images are utilized to transfer the knowledge to the student network~\cite{fang2019data, micaelli2019zero, Choi2020DataFreeNQ}.
Online adversarial feature map distillation (AFD)~\cite{chung2020feature}, meanwhile, has been proposed to utilize adversarial training to distill knowledge using two or three different networks at each feature extractor. As the generator of GAN, the feature extractors are trained to deceive the discriminators.
Xu and Li~\cite{XU2022242} proposed a contrastive adversarial KD with a designed two-stage framework: one is for feature distillation using adversarial and contrastive learning and another is for KD.

%-------------------------------------------------------------------------
\section{Preliminaries}
Before introducing our method, we formulate the EM distance and present some preliminaries.
There exist multiple metrics that can be used to quantify the dissimilarity or distance and diversity between two distinct probability distributions. Among the various measurements, KL divergence, Jenson-Shannon (JS) divergence, and Total variation (TV) distance are representative and widely employed. The EM distance has been adopted in WGAN-based approaches and is extensively utilized.

To explain the excerpts from the WGAN~\cite{arjovsky2017wasserstein}
descriptions, the distances, including the TV distance, KL divergence, JS divergence, and EM distance, are defined as follows:
%-------------------------------------------------------------------------
\begin{equation}
\begin{aligned}
    \delta({\mathbb{P}_r}, {\mathbb{P}_g})=\sup_{A\in\Sigma}|{\mathbb{P}_r}(A)-{\mathbb{P}_r}(A)|
\end{aligned}
\label{eq:TV_def}
\end{equation}
%-------------------------------------------------------------------------
\begin{equation}
\begin{aligned}
    KL({\mathbb{P}_r}||{\mathbb{P}_g})=\int\log(\frac{\mathbb{P}_r (x)}{\mathbb{P}_g (x)})\mathbb{P}_r(x)~d\mu(x)
\end{aligned}
\label{eq:KLD_def}
\end{equation}
%-------------------------------------------------------------------------
\begin{equation}
\begin{aligned}
    JS({\mathbb{P}_r},{\mathbb{P}_g})=KL({\mathbb{P}_r}||{\mathbb{P}_m})+KL({\mathbb{P}_r}||{\mathbb{P}_m})
\end{aligned}
\label{eq:JSD_def}
\end{equation}
%-------------------------------------------------------------------------
\begin{equation}
\begin{aligned}
    W({\mathbb{P}_r},{\mathbb{P}_g})=\sup_{\gamma\in\Pi({\mathbb{P}_r},{\mathbb{P}_g})}\mathbb{E}_{(x,y)\sim\gamma}[||x-y||]
\end{aligned}
\label{eq:EM_def}
\end{equation}

\noindent where $\mathbb{P}_r$ and $\mathbb{P}_g$ are indicated real data distribution and generated data distribution, respectively. $\Sigma$ represents the set of all the Borel subsets of $\mathfrak{X}$, and $\mathfrak{X}$ is a compact metric set. Also, $\mathbb{P}_m$ is equivalent with $(\mathbb{P}_r+\mathbb{P}_g)/2$, $\gamma(x,y)$ is a joint distribution that represents transportation quantity as a cost from $\mathbb{P}_r$ to $\mathbb{P}_g$.

However, this has a prerequisite that the distributions must overlap to some extent. More specifically, let $\mathbb{P}_0$ be a distribution of $(0, Z)\in R^2$ and $0$ is on the axis and $Z$ is on the other axis. These two axes are orthogonal. $\mathbb{P}_\theta$ is also a distribution with a single parameter $\theta$.
If the probabilities are equivalent to each other or do not overlap at all, the measured distance becomes a constant or infinity, indicating that the distance is useless. On the other hand, the distance from EM to $|\theta|$ is maintained and is independent of whether the distributions overlap or not. It is evident that, in the prerequisite, the equations are as follows
%-------------------------------------------------------------------------
\begin{equation}
\begin{aligned}
    \delta({\mathbb{P}_0}, {\mathbb{P}_\theta})=
    \begin{cases}
        1, &\text{if } \theta \neq 0\\
        0, &\text{if } \theta = 0
    \end{cases}
\end{aligned}
\label{eq:TV_cond}
\end{equation}
%-------------------------------------------------------------------------
\begin{equation}
\begin{aligned}
    KL({\mathbb{P}_0}||{\mathbb{P}_\theta})=KL({\mathbb{P}_\theta}||{\mathbb{P}_0})=
    \begin{cases}
        +\infty, &\text{if } \theta \neq 0\\
        0, &\text{if } \theta = 0
    \end{cases}
\end{aligned}
\label{eq:KLD_cond}
\end{equation}
%-------------------------------------------------------------------------
\begin{equation}
\begin{aligned}
    JS({\mathbb{P}_0},{\mathbb{P}_\theta})=
    \begin{cases}
        \log 2, &\text{if } \theta \neq 0\\
        0, &\text{if } \theta = 0
    \end{cases}    
\end{aligned}
\label{eq:JSD_cond}
\end{equation}
%-------------------------------------------------------------------------
\begin{equation}
\begin{aligned}
    W({\mathbb{P}_0},{\mathbb{P}_\theta})=|\theta|
\end{aligned}
\label{eq:EM_cond}
\end{equation}

\noindent
Our approach, AI-KD, aims to align the distributions between the pre-trained and student models, not directly mimicking the pre-trained model. In AFD,~\cite{chung2020feature}
the discriminator is used to generate a probability for predicting the origin of inputs from certain models. However, in WGAN, the discriminator exploits a scalar value through the EM distance, which enables more effective gradient delivery, resulting in improved learning for both the discriminator and the generator. For these reasons, AI-KD has adopted partially the WGAN scheme to leverage the EM distance.

%-------------------------------------------------------------------------
\section{Methodology}
\label{sec:method}
%-------------------------------------------------------------------------
\begin{figure*}[t]
    \centering
    \resizebox{0.95\linewidth}{!}{
    \setlength{\tabcolsep}{5pt}
    \begin{tabular}{c}
        \includegraphics[width=1\linewidth]{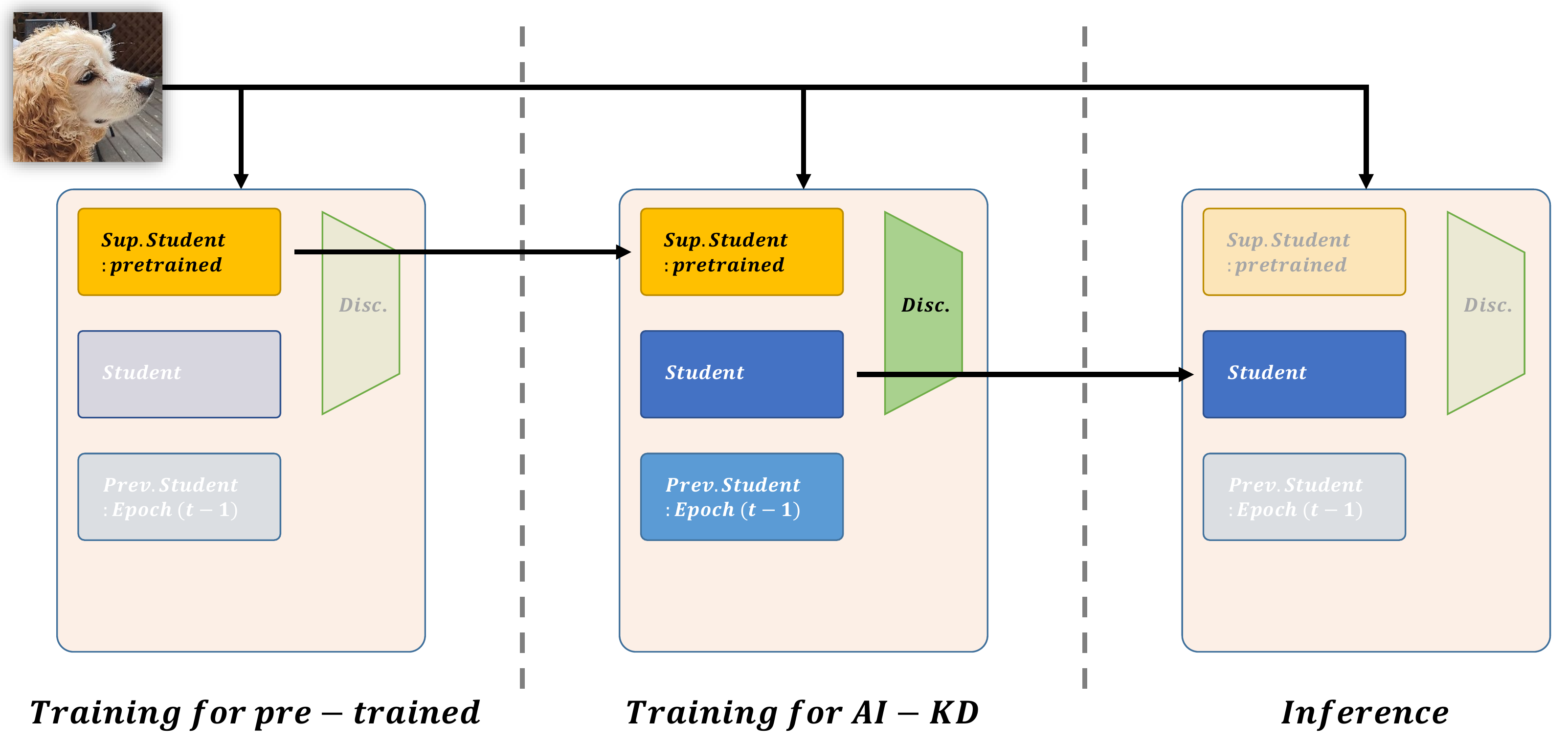}
    \end{tabular}}
    \caption{The overall framework of AI-KD. Training models through AI-KD have a two-step process. The first phase is training the pre-trained model from scratch. The pre-trained model serves as the baseline and is utilized as the superior student model for the next phase. Then the second phase is training the student model via AI-KD. The last phase is for inference, during which the student model is utilized without the need for the superior student model, the previous student model, or the discriminator.}
    \label{fig:train_inference}
\end{figure*}
%-------------------------------------------------------------------------
\subsection{Overall framework}
AI-KD utilizes three same-size network architectures but different objectives, which are the pre-trained model, the student model, and the previous student model.
As shown in Figure~\ref{fig:train_inference},
AI-KD requires two separate, but sequential training procedures, which are one for the pre-trained model and another for our proposed method.
The first step is acquiring a pre-trained model. The model architecture and training parameters (e.g., learning rate, weight decay, batch size, and standard data augmentation approaches) of the pre-trained model are identical to the student model. 
The acquired pre-trained model is utilized as named superior student model to guide mimicking output distributions of the student model to the pre-trained model in the next phase.
Then the second step is training the student model to utilize two different models, which are the pre-trained model and the old student model from the previous epoch. In the inference phase, the pre-trained and previous student models are not activated, and the well-trained student model is only utilized.

The discriminator aims to distinguish whether the output distributions are from the student model or the pre-trained model. For that reason, the discriminator requires as input one-dimensional and consists of stacking two different sizes of fully-connected layers. The activation function of the discriminator is Leaky ReLU. The single one-dimensional batch normalization technique is employed to better utilize the discriminator. The batch normalization and the activation function are placed between fully-connected layers as Fully connected layer-1D Batch Normalization-LeakyReLU-Fully connected layer. The discriminator is employed only in the second phase and is not utilized in the acquiring pre-trained model phase and the inference phase.
%-------------------------------------------------------------------------
\begin{figure*}[!t]
    \centering
    \resizebox{0.9\linewidth}{!}{
    \setlength{\tabcolsep}{5pt}
    \begin{tabular}{c}
        \includegraphics[width=1.\linewidth]{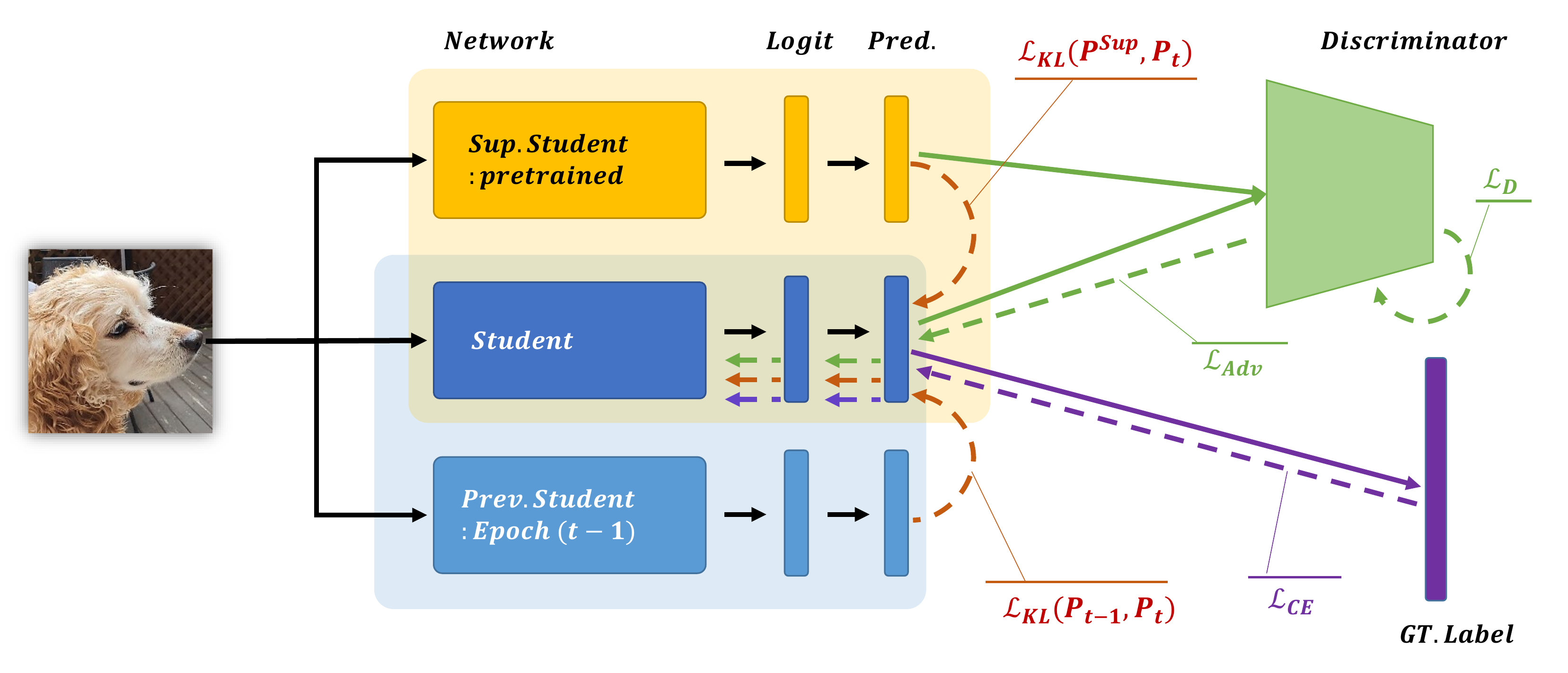}
    \end{tabular}}
    \caption{The overall losses of AI-KD. The student model is trained implicitly distilled knowledge from the superior pre-trained and previous student models. Through adversarial learning, the student is aligned to the distributions of the superior pre-trained model. The solid lines are feed-forward operations, and the dashed lines represent backward operations used to update the model from each loss.}
    \label{fig:overview}
\end{figure*}
%-------------------------------------------------------------------------
\subsection{Objective of learning via AI-KD losses}
In this work, we focus on fully-supervised classification tasks. The input data and the corresponding ground-truth labels are denoted as $x\in X$ and $y\in Y=\{1,\dots ,C\}$, respectively. Given a student model $S$, $S$ outputs the posterior predictive distribution of each input using a softmax classifier.
%-------------------------------------------------------------------------
\begin{equation}
\begin{aligned}
    p_{i}(x;\tau) = \frac{exp(z_{i}(x)/\tau)}{\sum_{i=1}^{C}exp(z_{i}(x)/\tau)}
\end{aligned}
\label{eq:softmax}
\end{equation}
%-------------------------------------------------------------------------
\noindent where $z_i$ is the logit of the model for class $i$, and $\tau > 0$ is the temperature scaling parameter to soften $p_i$ for better distillation. 

In conventional knowledge distillation, the loss for training the student model is given by
%-------------------------------------------------------------------------
\begin{equation}
\begin{aligned}
    \mathcal{L}_{KD} = (1-\alpha)\mathcal{H}(q,p^S(x))+\alpha~{\tau}^{2}~\mathcal{H}(p^T(x;\tau),p^S(x;\tau))
\end{aligned}
\label{eq:kdloss}
\end{equation}
%-------------------------------------------------------------------------
\noindent where $\mathcal{H}(\cdot ,\cdot)$ indicates the standard cross-entropy, $\alpha$ is the balancing hyper-parameter, and $p^S$ and $p^T$ are output probabilities of a student and teacher network, respectively. $q$ denotes $C$-dimensional one-hot of $y$.

Let $D_{KL}$ denotes the KL divergence, and $\mathcal{H}(p^T, p^S)=D_{KL}(p^T, p^S)-\mathcal{H}(p^T)$, then we can reformulate Equation~\ref{eq:kdloss} as
%-------------------------------------------------------------------------
\begin{equation}
\begin{aligned}
    \mathcal{L}_{KD} &= (1-\alpha)\mathcal{H}(y,p^S(x))+\alpha~{\tau}^{2}~D_{KL}(p^{T}(x;\tau),p^S(x;\tau))-\alpha~{\tau}^{2}~\mathcal{H}(p^{T}(x;\tau))\\
    &= (1-\alpha)\mathcal{H}(y,p^S(x))+\alpha~{\tau}^{2}~D_{KL}(p^{T}(x;\tau),p^S(x;\tau))     
\end{aligned}
\label{eq:kdloss_2}
\end{equation}
%-------------------------------------------------------------------------
\noindent where $p^T$ from the pre-trained model should have deterministic probability distributions, then the entropy $\mathcal{H}(\cdot)$ is treated as a constant for the fixed model and can be ignored.

%-------------------------------------------------------------------------
From Equation~\ref{eq:kdloss_2}, we separate the $D_{KL}$ into two terms for distilling deterministic and progressive knowledge from different models.
And, as depicted in Figure~\ref{fig:overview}, the overall architecture of AI-KD consists of two different methods: a deterministic and progressive knowledge distillation method and an adversarial learning method to align distributions between pre-trained and student models.
The student network learns different knowledge using two homogeneous networks of deterministic probability distributions, which are the pre-trained and the previous epoch model.
We define three different networks: a superior pre-trained student model $S^{Sup}$, a student model $S_{t}$, and a previous epoch $(t-1)^{th}$ student model $S_{t-1}$.
The output of the superior pre-trained model, $p^{Sup}$, and the output of the previous epoch student model, $p_{t-1}$, are used to distill implicit knowledge to the student model by penalizing the output of the student model, $p_t$.

The model is not from the ImageNet dataset; instead is the baseline of each dataset and is trained from scratch following the standard training parameters.
The role of $S^{Sup}$ is that as a good guide and teacher, the superior model guides the student model to improve performance. In this sense, the loss function for distillation of $S^{Sup}$ is named guide, $\mathcal{L}_{G}$ and aims that $S$ directly mimics the logits of $S^{Sup}$. $\mathcal{L}_{G}$ can be written as
%-------------------------------------------------------------------------
\begin{equation}
\begin{aligned}
    \mathcal{L}_{G} = {\tau_{G}}^{2}~D_{KL}(p^{Sup}(x;\tau_{G}),~p_{t}(x;\tau_{G}))
\end{aligned}
\label{eq:tfkd}
\end{equation}
%-------------------------------------------------------------------------
\indent 
$L_G$ may induce rapid weight variation in epochs, leading to degraded performance.
To prevent performance degradation and unsteady learning, the student model is available to acquire progressive information from previous knowledge.
We empirically observed that the $t-1^{th}$ epoch model could provide effective knowledge distilled for training progressively. The loss function of learning from previous predictions is coined $L_P$ and affects regularization by logit smoothing. In this manner, $L_P$ can be written as
%-------------------------------------------------------------------------
\begin{equation}
\begin{aligned}
    \mathcal{L}_{P} = {\tau_{P}}^{2}~D_{KL}(p_{t-1}(x;\tau_{P}),~p_{t}(x;\tau_{P}))
\end{aligned}
\label{eq:pskd}
\end{equation}
%-------------------------------------------------------------------------
\indent
Although the student model is regularized by $L_G$ and $L_P$, the predictive logits of the student model can just be overfitting to the logits of the pre-trained model. To prevent overfitting, AI-KD employs an adversarial learning method utilizing a discriminator $D$ that is composed of two fully-connected layers to distinguish $p^{Sup}$ and $p_t$. The goal of the proposed method is to align the predictive distributions of the student model to the predictive distributions of the pre-trained model. By adopting the adversarial learning method, the student model is trained to fool the discriminator. A well-trained discriminator helps the student model to be trained to generalize the predictive distributions to the pre-trained model.
Motivated by WGAN-GP~\cite{gulrajani2017improved}, $\mathcal{L}_{D}$ is defined as a critic loss that scores between the pre-trained model and the student model.
%-------------------------------------------------------------------------
\begin{equation}
\begin{aligned}
    \mathcal{L}_{D} = \mathbb{E}_{x\sim p_t(x)}[D(S_t(x))] - \mathbb{E}_{x\sim p^{Sup}(x)}[D(S^{Sup}(x))] + \lambda\mathbb{E}_{\hat{x}\sim p_{\hat{x}}}[(\vert\vert\nabla_{\hat{x}}D(\hat{x})\vert\vert_{2}-1)^2]
\end{aligned}
\label{eq:disc}
\end{equation}
%-------------------------------------------------------------------------
where the last term indicates gradient penalty, and the middle term is ignored which is from the deterministic.
In this manner, the adversarial loss $\mathcal{L}_{A}$ for the student model should be contrastive to $\mathcal{L}_{D}$ and aims to regularize the predictive distributions of the student model to align with the output distributions of the pre-trained model. $\mathcal{L}_{A}$ can be written as
%-------------------------------------------------------------------------
\begin{equation}
\begin{aligned}
    \mathcal{L}_{A} = -\mathbb{E}_{x\sim p_t(x)}[D(S_t(x))]
\end{aligned}
\label{eq:adv}
\end{equation}

%-------------------------------------------------------------------------
In conclusion, the AI-KD loss consists of two different losses, one loss is for training the discriminator as follows to Equation~\ref{eq:disc}, and another loss is for training the student model, $\mathcal{L}_{AI}$ is described as follows:
%-------------------------------------------------------------------------
\begin{equation}
\begin{aligned}
    \mathcal{L}_{AI} = (1 - {\alpha}_{P} - {\alpha}_{G})~\mathcal{H}(q, p_t) + {\alpha}_{P}~\mathcal{L}_{P} + {\alpha}_{G}~\mathcal{L}_{G} + \omega~\mathcal{L}_{A}\\
\end{aligned}
\label{eq:tot}
\end{equation}
%-------------------------------------------------------------------------
where the coefficients $\alpha_{P}$, $\alpha_{G}$, and $\omega$ are deployed to balance the cross-entropy, KD losses by different models, and adversarial loss, respectively.

To sum up the objective of each proposed loss, the $\mathcal{L}_{G}$ guides that $S_t$ directly mimics the logits of $S^{Sup}$, may induce rapid weight variation in epochs, and leads to degraded performance. $\mathcal{L}_{P}$ affects to regularize as logit smoothing by distilling information from the previous epoch to prevent degradation. However, the overfitting issue is still inherent. To address the issue, $\mathcal{L}_{A}$ is the crucial loss function that the predictive distributions of $S_t$ lead to aligning with the output distributions of $S^{Sup}$.

%-------------------------------------------------------------------------
\section{Experiment results}
\label{sec:experimentalresults}
In this section, we describe the effectiveness of our proposed method on the image classification task with network architectures on various coarse and fine-grained datasets.
In Section~\ref{sec:exp_data_metric}, we first introduce the datasets and the metrics utilized to verify the performance of our proposed method. And Section~\ref{sec:exp_implementation} describes specific training environments and parameters. Then we confirm to compare the classification results with logit-based Self-KD representative methods, in Section~\ref{sec:exp_classification}. For a fair comparison, the hyper-parameters of comparison methods, such as CS-KD~\cite{Yun_2020_CVPR}, TF-KD~\cite{yuan2020revisiting},
PS-KD~\cite{Kim_2021_ICCV}, TF-FD~\cite{li2022self}, and ZipfsLS~\cite{liang2022efficient} are set as reported.
All the reported performances are average results over three runs of each experiment. In Section~\ref{sec:exp_ablation}, we experiment with ablation studies and empirical analyses to confirm the effects of losses and balance weight parameters for each loss. Lastly, we evaluate compatibility with other data augmentation techniques to confirm that AI-KD is a valid regularization method and could be employed with other approaches independently, in Section~\ref{sec:exp_data_augmentations}.
%-------------------------------------------------------------------------
\begin{table}[t]
\begin{center}
\footnotesize
\setlength{\tabcolsep}{1.0pt}
\begin{tabular}{p{0.3\linewidth}p{0.25\linewidth}P{0.2\linewidth}P{0.1\linewidth}P{0.1\linewidth}}
    \toprule
    \multirow{2}{*}{Category} &\multirow{2}{*}{Dataset} &\multirow{2}{*}{Classes} &\multicolumn{2}{c}{Images}\\
    \cmidrule{4-5}
    & & &Train &Valid\\
    \midrule
    \midrule
    \multirow{2}{*}{Coarse} &CIFAR-100~\cite{krizhevsky2009learning} &10 &50,000 &10,000\\
    &Tiny-ImageNet~\cite{krizhevsky2009learning} &200 &100,000 &10,000\\
    \midrule
    \multirow{4}{*}{Fine-grained} &CUB200-2011~\cite{wah2011caltech} &200 &5,994 &5,794\\
    &Stanford Dogs~\cite{khosla2011novel} &120 &12,000 &8,580\\
    &MIT67~\cite{quattoni2009recognizing} &67 &5,360 &1,340\\
    &FGVC aircraft~\cite{maji2013fgvcair} &100 &6,667 &3,334\\
    \bottomrule
\end{tabular}
\end{center}
\caption{The datasets to evaluate classification tasks. The coarse datasets consist of classifying interspecies, while fine-grained datasets are composed of recognizing specific species.}
\label{tab:dataset}
\end{table}

%-------------------------------------------------------------------------
\subsection{Datasets and evaluation metrics}
\label{sec:exp_data_metric}
\paragraph{Datasets}
We evaluate the proposed method for image classification performance using a variety of coarse and fine-grained datasets, as shown in Table~\ref{tab:dataset}. Fine-grained datasets are composed for recognizing specific species, e.g., canines, indoor scenes, vehicles, and birds. While coarse datasets are intended to evaluate classifying interspecies objects.
The proposed method is evaluated on coarse datasets, CIFAR-100~\cite{krizhevsky2009learning} and Tiny-ImageNet\footnote[1]{https://tiny-imagenet.herokuapp.com}, using PreAct ResNet-18~\cite{he2016identity}, PreAct ResNet-50, and DenseNet-121~\cite{huang2017densely}, and on fine-grained datasets, CUB200-2011~\cite{wah2011caltech}, Stanford Dogs~\cite{khosla2011novel}, MIT67~\cite{quattoni2009recognizing}, and FGVC aircraft~\cite{maji2013fgvcair}, using ResNet-18~\cite{he2016deep}, ResNet-50, and DenseNet-121.
%-------------------------------------------------------------------------
\paragraph{Metrics}
The Top-$\mathcal{K}$ error rate is the fraction of evaluation samples that from the logit results, the correct label does not include in the Top-$\mathcal{K}$ confidences. We utilize the Top-1 and Top-5 error rates to evaluate the regularization performance of AI-KD. F1-score is described as $\frac{2 P R}{P+R}$, where $\mathcal{P}$ and $\mathcal{R}$ indicate precision and recall. The metric represents the balance between precision and recall and the score is widely used for comparing imbalanced datasets. In addition, we measure calibration effects using expected calibration error~(ECE)~\cite{naeini2015obtaining, guo2017calibration}. ECE approximates the expected difference between the average confidence and the accuracy.

%-------------------------------------------------------------------------
\subsection{Implementation details}
\label{sec:exp_implementation}
All experiments were performed on NVIDIA RTX 2080 Ti system with PyTorch. We utilize the standard augmentation techniques, such as random crop after padding and random horizontal flip. All the experiments are trained for 300 epochs using SGD with the Nesterov momentum and weight decay, 0.9 and 0.0005, respectively. The initial learning rate is 0.1 and is decayed by a factor of 0.1 at epochs 150 and 225. We utilize different batch sizes depending on the dataset type, 128 for the coarse datasets and 32 for the fine-grained datasets.

In the proposed AI-KD, the goal of the discriminator network is to distinguish the outputs between the superior pre-trained model and the student model. In this manner, the student model is trained to mimic the superior model, and the probability distributions of the student model also try to align with the probability distributions of the superior model. The discriminative network requires an input of one-dimensional simple information that is the logits from the student and the superior models. In this sense, we propose a light series of fully connected (FC) layer-1D Batch Normalization-LeakyReLU-FC layer. These two FC layers have different dimensions and are denoted as $FC_1$ and $FC_2$. The layers have 64 and 32 output features, respectively. The optimizer of the discriminator is Adam and the initial learning rate is 0.0001. For gradient penalty, we utilize that the weight is set to 10. 
%-------------------------------------------------------------------------
\begin{table*}[!t]
\begin{center}
\footnotesize
\setlength{\tabcolsep}{1.0pt}
\resizebox{\linewidth}{!}{
\begin{tabular}{P{0.09\linewidth}P{0.1\linewidth}p{0.11\linewidth}P{0.125\linewidth}P{0.125\linewidth}P{0.125\linewidth}P{0.125\linewidth}P{0.125\linewidth}P{0.125\linewidth}}
    \toprule
    \multirow{2}{*}[-5pt]{Metric} &\multirow{2}{*}[-5pt]{Model} &\multirow{2}{*}[-5pt]{Method} &\multicolumn{6}{c}{Dataset}\\
    \cmidrule{4-9}
    & & &CIFAR-100 &\specialcell[c]{Tiny\\ImageNet} &\specialcell[c]{CUB\\200-2011} &\specialcell[c]{Stanford\\Dogs} &MIT67 &\specialcell[c]{FGVC\\Aircraft}\\
    \midrule
    \midrule
    \multirow{12}{*}[-2pt]{\shortstack{Top-1\\Error}} &\multirow{7}{*}[-2pt]{\shortstack{ResNet\\18}} &Baseline &23.97\fontsize{0.2cm}{0.2cm}\selectfont$\pm$0.29 &46.28\fontsize{0.2cm}{0.2cm}\selectfont$\pm$0.19 &36.65\fontsize{0.2cm}{0.2cm}\selectfont$\pm$0.87 &32.61\fontsize{0.2cm}{0.2cm}\selectfont$\pm$0.55 &39.90\fontsize{0.2cm}{0.2cm}\selectfont$\pm$0.16 &18.61\fontsize{0.2cm}{0.2cm}\selectfont$\pm$1.16\\
    & &CS-KD~\cite{Yun_2020_CVPR} &21.67\fontsize{0.2cm}{0.2cm}\selectfont$\pm$0.25 &\textbf{41.68\fontsize{0.2cm}{0.2cm}\selectfont$\pm$0.15} &\underline{33.93\fontsize{0.2cm}{0.2cm}\selectfont$\pm$0.29} &30.87\fontsize{0.2cm}{0.2cm}\selectfont$\pm$0.05 &41.42\fontsize{0.2cm}{0.2cm}\selectfont$\pm$0.45 &19.66\fontsize{0.2cm}{0.2cm}\selectfont$\pm$0.02\\
    & &TF-KD~\cite{yuan2020revisiting} &22.07\fontsize{0.2cm}{0.2cm}\selectfont$\pm$0.38 &43.66\fontsize{0.2cm}{0.2cm}\selectfont$\pm$0.08 &35.92\fontsize{0.2cm}{0.2cm}\selectfont$\pm$0.58 &31.25\fontsize{0.2cm}{0.2cm}\selectfont$\pm$0.27 &38.08\fontsize{0.2cm}{0.2cm}\selectfont$\pm$0.55 &19.27\fontsize{0.2cm}{0.2cm}\selectfont$\pm$0.49\\
    & &PS-KD~\cite{Kim_2021_ICCV} &\underline{21.16\fontsize{0.2cm}{0.2cm}\selectfont$\pm$0.22} &43.45\fontsize{0.2cm}{0.2cm}\selectfont$\pm$0.28 &34.01\fontsize{0.2cm}{0.2cm}\selectfont$\pm$0.74 &\underline{29.20\fontsize{0.2cm}{0.2cm}\selectfont$\pm$0.23} &\underline{37.55\fontsize{0.2cm}{0.2cm}\selectfont$\pm$0.61} &\underline{16.80\fontsize{0.2cm}{0.2cm}\selectfont$\pm$0.18}\\
    & &TF-FD~\cite{li2022self}
    &23.77\fontsize{0.2cm}{0.2cm}\selectfont$\pm$0.09 &45.95\fontsize{0.2cm}{0.2cm}\selectfont$\pm$0.17 &34.62\fontsize{0.2cm}{0.2cm}\selectfont$\pm$0.63 &33.09\fontsize{0.2cm}{0.2cm}\selectfont$\pm$0.11 &38.06\fontsize{0.2cm}{0.2cm}\selectfont$\pm$0.63 &16.65\fontsize{0.2cm}{0.2cm}\selectfont$\pm$0.08\\
    & &ZipfsLS~\cite{liang2022efficient}
    &22.17\fontsize{0.2cm}{0.2cm}\selectfont$\pm$0.28 &42.21\fontsize{0.2cm}{0.2cm}\selectfont$\pm$0.12 &37.12\fontsize{0.2cm}{0.2cm}\selectfont$\pm$0.71 &34.68\fontsize{0.2cm}{0.2cm}\selectfont$\pm$0.20 &42.62\fontsize{0.2cm}{0.2cm}\selectfont$\pm$0.51 &21.34\fontsize{0.2cm}{0.2cm}\selectfont$\pm$0.36\\
    \cmidrule{3-9}
    & &\textbf{AI-KD} &\textbf{19.87\fontsize{0.2cm}{0.2cm}\selectfont$\pm$0.10} &\underline{41.84\fontsize{0.2cm}{0.2cm}\selectfont$\pm$0.10} &\textbf{29.59\fontsize{0.2cm}{0.2cm}\selectfont$\pm$0.27} &\textbf{28.61\fontsize{0.2cm}{0.2cm}\selectfont$\pm$0.21} &\textbf{36.84\fontsize{0.2cm}{0.2cm}\selectfont$\pm$0.34} &\textbf{15.43\fontsize{0.2cm}{0.2cm}\selectfont$\pm$0.15}\\
    \cmidrule{2-9}
    &\multirow{7}{*}[-2pt]{\shortstack{ResNet\\50}} &Baseline
    &21.57\fontsize{0.2cm}{0.2cm}\selectfont$\pm$0.32 &41.73\fontsize{0.2cm}{0.2cm}\selectfont$\pm$0.12 &37.53\fontsize{0.2cm}{0.2cm}\selectfont$\pm$1.28 &32.84\fontsize{0.2cm}{0.2cm}\selectfont$\pm$0.42 &38.83\fontsize{0.2cm}{0.2cm}\selectfont$\pm$0.15 &17.94\fontsize{0.2cm}{0.2cm}\selectfont$\pm$1.18\\
    & &CS-KD~\cite{Yun_2020_CVPR}
    &21.33\fontsize{0.2cm}{0.2cm}\selectfont$\pm$0.13 &39.17\fontsize{0.2cm}{0.2cm}\selectfont$\pm$0.41 &40.26\fontsize{0.2cm}{0.2cm}\selectfont$\pm$0.25 &35.19\fontsize{0.2cm}{0.2cm}\selectfont$\pm$1.81 &47.80\fontsize{0.2cm}{0.2cm}\selectfont$\pm$0.58 &26.57\fontsize{0.2cm}{0.2cm}\selectfont$\pm$1.97\\
    & &TF-KD~\cite{yuan2020revisiting}
    &20.07\fontsize{0.2cm}{0.2cm}\selectfont$\pm$0.22 &39.69\fontsize{0.2cm}{0.2cm}\selectfont$\pm$0.35 &\underline{37.02\fontsize{0.2cm}{0.2cm}\selectfont$\pm$2.88} &\textbf{28.71\fontsize{0.2cm}{0.2cm}\selectfont$\pm$0.42} &\underline{36.60\fontsize{0.2cm}{0.2cm}\selectfont$\pm$0.26} &\underline{16.12\fontsize{0.2cm}{0.2cm}\selectfont$\pm$0.61}\\
    & &PS-KD~\cite{Kim_2021_ICCV}
    &\underline{19.93\fontsize{0.2cm}{0.2cm}\selectfont$\pm$0.88} &\underline{39.08\fontsize{0.2cm}{0.2cm}\selectfont$\pm$0.15} 
    &39.35\fontsize{0.2cm}{0.2cm}\selectfont$\pm$1.90 &29.14\fontsize{0.2cm}{0.2cm}\selectfont$\pm$1.95 
    &39.74\fontsize{0.2cm}{0.2cm}\selectfont$\pm$0.79 &19.41\fontsize{0.2cm}{0.2cm}\selectfont$\pm$1.25\\
    & &TF-FD~\cite{li2022self}
    &20.69\fontsize{0.2cm}{0.2cm}\selectfont$\pm$0.12 &39.81\fontsize{0.2cm}{0.2cm}\selectfont$\pm$0.47 &38.16\fontsize{0.2cm}{0.2cm}\selectfont$\pm$0.56 &32.04\fontsize{0.2cm}{0.2cm}\selectfont$\pm$0.11 &39.89\fontsize{0.2cm}{0.2cm}\selectfont$\pm$0.05 &18.24\fontsize{0.2cm}{0.2cm}\selectfont$\pm$1.18\\
    & &ZipfsLS~\cite{liang2022efficient}
    &21.89\fontsize{0.2cm}{0.2cm}\selectfont$\pm$0.12 &39.70\fontsize{0.2cm}{0.2cm}\selectfont$\pm$0.53 &43.71\fontsize{0.2cm}{0.2cm}\selectfont$\pm$0.34 &37.52\fontsize{0.2cm}{0.2cm}\selectfont$\pm$1.73 &49.39\fontsize{0.2cm}{0.2cm}\selectfont$\pm$0.37 &27.12\fontsize{0.2cm}{0.2cm}\selectfont$\pm$1.52\\
    \cmidrule{3-9}
    & &\textbf{AI-KD}
    &\textbf{19.42\fontsize{0.2cm}{0.2cm}\selectfont$\pm$0.59} &\textbf{38.79\fontsize{0.2cm}{0.2cm}\selectfont$\pm$0.02} &\textbf{32.68\fontsize{0.2cm}{0.2cm}\selectfont$\pm$0.33} &\underline{29.11\fontsize{0.2cm}{0.2cm}\selectfont$\pm$0.08} &\textbf{36.42\fontsize{0.2cm}{0.2cm}\selectfont$\pm$0.87} &\textbf{13.79\fontsize{0.2cm}{0.2cm}\selectfont$\pm$0.44}\\
    \midrule
    \multirow{12}{*}{\shortstack{F1 Score\\Macro}} &\multirow{7}{*}[-2pt]{\shortstack{ResNet\\18}} &Baseline &0.759\fontsize{0.2cm}{0.2cm}\selectfont$\pm$0.003 &0.536\fontsize{0.2cm}{0.2cm}\selectfont$\pm$0.002 &0.631\fontsize{0.2cm}{0.2cm}\selectfont$\pm$0.008 &0.663\fontsize{0.2cm}{0.2cm}\selectfont$\pm$0.006 &0.598\fontsize{0.2cm}{0.2cm}\selectfont$\pm$0.004 &0.821\fontsize{0.2cm}{0.2cm}\selectfont$\pm$0.002\\
    & &CS-KD~\cite{Yun_2020_CVPR} &0.781\fontsize{0.2cm}{0.2cm}\selectfont$\pm$0.002 &\textbf{0.581\fontsize{0.2cm}{0.2cm}\selectfont$\pm$0.002} &0.651\fontsize{0.2cm}{0.2cm}\selectfont$\pm$0.005 &0.681\fontsize{0.2cm}{0.2cm}\selectfont$\pm$0.001 &0.583\fontsize{0.2cm}{0.2cm}\selectfont$\pm$0.002 &0.804\fontsize{0.2cm}{0.2cm}\selectfont$\pm$0.000\\
    & &TF-KD~\cite{yuan2020revisiting} &0.779\fontsize{0.2cm}{0.2cm}\selectfont$\pm$0.004 &0.563\fontsize{0.2cm}{0.2cm}\selectfont$\pm$0.001 &0.638\fontsize{0.2cm}{0.2cm}\selectfont$\pm$0.006 &0.678\fontsize{0.2cm}{0.2cm}\selectfont$\pm$0.003 &0.621\fontsize{0.2cm}{0.2cm}\selectfont$\pm$0.007 &0.807\fontsize{0.2cm}{0.2cm}\selectfont$\pm$0.005\\
    & &PS-KD~\cite{Kim_2021_ICCV} &\underline{0.788\fontsize{0.2cm}{0.2cm}\selectfont$\pm$0.002} &0.560\fontsize{0.2cm}{0.2cm}\selectfont$\pm$0.003 &\underline{0.657\fontsize{0.2cm}{0.2cm}\selectfont$\pm$0.007} &\underline{0.697\fontsize{0.2cm}{0.2cm}\selectfont$\pm$0.002} &\underline{0.623\fontsize{0.2cm}{0.2cm}\selectfont$\pm$0.004} &\underline{0.831\fontsize{0.2cm}{0.2cm}\selectfont$\pm$0.002}\\
    & &TF-FD~\cite{li2022self}
    &0.764\fontsize{0.2cm}{0.2cm}\selectfont$\pm$0.009 &0.541\fontsize{0.2cm}{0.2cm}\selectfont$\pm$0.003 &0.652\fontsize{0.2cm}{0.2cm}\selectfont$\pm$0.006 &0.658\fontsize{0.2cm}{0.2cm}\selectfont$\pm$0.004 &0.616\fontsize{0.2cm}{0.2cm}\selectfont$\pm$0.007 &0.834\fontsize{0.2cm}{0.2cm}\selectfont$\pm$0.001\\
    & &ZipfsLS~\cite{liang2022efficient}
    &0.776\fontsize{0.2cm}{0.2cm}\selectfont$\pm$0.005 &0.575\fontsize{0.2cm}{0.2cm}\selectfont$\pm$0.003 &0.619\fontsize{0.2cm}{0.2cm}\selectfont$\pm$0.004 &0.652\fontsize{0.2cm}{0.2cm}\selectfont$\pm$0.004 &0.571\fontsize{0.2cm}{0.2cm}\selectfont$\pm$0.006 &0.782\fontsize{0.2cm}{0.2cm}\selectfont$\pm$0.001\\
    \cmidrule{3-9}
    & &\textbf{AI-KD} &\textbf{0.801\fontsize{0.2cm}{0.2cm}\selectfont$\pm$0.001} &\textbf{0.581\fontsize{0.2cm}{0.2cm}\selectfont$\pm$0.002} &\textbf{0.701\fontsize{0.2cm}{0.2cm}\selectfont$\pm$0.003} &\textbf{0.703\fontsize{0.2cm}{0.2cm}\selectfont$\pm$0.002} &\textbf{0.624\fontsize{0.2cm}{0.2cm}\selectfont$\pm$0.004} &\textbf{0.845\fontsize{0.2cm}{0.2cm}\selectfont$\pm$0.001}\\
    \cmidrule{2-9}
    &\multirow{7}{*}[-2pt]{\shortstack{ResNet\\50}} &Baseline 
    &0.784\fontsize{0.2cm}{0.2cm}\selectfont$\pm$0.003 &0.582\fontsize{0.2cm}{0.2cm}\selectfont$\pm$0.001 
    &0.624\fontsize{0.2cm}{0.2cm}\selectfont$\pm$0.011 &0.662\fontsize{0.2cm}{0.2cm}\selectfont$\pm$0.003 
    &0.609\fontsize{0.2cm}{0.2cm}\selectfont$\pm$0.001 &0.821\fontsize{0.2cm}{0.2cm}\selectfont$\pm$0.012\\
    & &CS-KD~\cite{Yun_2020_CVPR} 
    &0.787\fontsize{0.2cm}{0.2cm}\selectfont$\pm$0.001 &\underline{0.607\fontsize{0.2cm}{0.2cm}\selectfont$\pm$0.004} &0.595\fontsize{0.2cm}{0.2cm}\selectfont$\pm$0.003 &0.635\fontsize{0.2cm}{0.2cm}\selectfont$\pm$0.011 &0.520\fontsize{0.2cm}{0.2cm}\selectfont$\pm$0.004 &0.734\fontsize{0.2cm}{0.2cm}\selectfont$\pm$0.009\\
    & &TF-KD~\cite{yuan2020revisiting} 
    &\underline{0.798\fontsize{0.2cm}{0.2cm}\selectfont$\pm$0.002} &0.599\fontsize{0.2cm}{0.2cm}\selectfont$\pm$0.001 &\underline{0.626\fontsize{0.2cm}{0.2cm}\selectfont$\pm$0.030} &\textbf{0.708\fontsize{0.2cm}{0.2cm}\selectfont$\pm$0.004} &\underline{0.629\fontsize{0.2cm}{0.2cm}\selectfont$\pm$0.004} &\underline{0.839\fontsize{0.2cm}{0.2cm}\selectfont$\pm$0.006}\\
    & &PS-KD~\cite{Kim_2021_ICCV}
    &0.797\fontsize{0.2cm}{0.2cm}\selectfont$\pm$0.010 &0.606\fontsize{0.2cm}{0.2cm}\selectfont$\pm$0.001 
    &0.603\fontsize{0.2cm}{0.2cm}\selectfont$\pm$0.019 &0.698\fontsize{0.2cm}{0.2cm}\selectfont$\pm$0.020 
    &0.598\fontsize{0.2cm}{0.2cm}\selectfont$\pm$0.007 &0.805\fontsize{0.2cm}{0.2cm}\selectfont$\pm$0.013\\
    & &TF-FD~\cite{li2022self} 
    &0.793\fontsize{0.2cm}{0.2cm}\selectfont$\pm$0.002 &0.601\fontsize{0.2cm}{0.2cm}\selectfont$\pm$0.005
    &0.616\fontsize{0.2cm}{0.2cm}\selectfont$\pm$0.006 &0.670\fontsize{0.2cm}{0.2cm}\selectfont$\pm$0.001 
    &0.595\fontsize{0.2cm}{0.2cm}\selectfont$\pm$0.002 &0.826\fontsize{0.2cm}{0.2cm}\selectfont$\pm$0.022\\
    & &ZipfsLS~\cite{liang2022efficient} 
    &0.780\fontsize{0.2cm}{0.2cm}\selectfont$\pm$0.001 &0.601\fontsize{0.2cm}{0.2cm}\selectfont$\pm$0.005 
    &0.580\fontsize{0.2cm}{0.2cm}\selectfont$\pm$0.004 &0.611\fontsize{0.2cm}{0.2cm}\selectfont$\pm$0.020 
    &0.507\fontsize{0.2cm}{0.2cm}\selectfont$\pm$0.001 &0.730\fontsize{0.2cm}{0.2cm}\selectfont$\pm$0.018\\
    \cmidrule{3-9}
    & &\textbf{AI-KD} 
    &\textbf{0.806\fontsize{0.2cm}{0.2cm}\selectfont$\pm$0.006} &\textbf{0.612\fontsize{0.2cm}{0.2cm}\selectfont$\pm$0.001} &\textbf{0.671\fontsize{0.2cm}{0.2cm}\selectfont$\pm$0.003} &\underline{0.699\fontsize{0.2cm}{0.2cm}\selectfont$\pm$0.001} &\textbf{0.632\fontsize{0.2cm}{0.2cm}\selectfont$\pm$0.010} &\textbf{0.862\fontsize{0.2cm}{0.2cm}\selectfont$\pm$0.004}\\
    \bottomrule
\end{tabular}}
\end{center}
\vspace*{-3mm}
\caption{Image classification Top-1 error and F1 Score results on ResNet-18 and ResNet-50 with various datasets. We report mean and standard deviation over three runs. The best and second-best results are indicated in bold and underlined, respectively.}
\vspace*{-3mm}
\label{tab:coarse_t1}
\end{table*}
%-------------------------------------------------------------------------
\begin{table*}[!ht]
\begin{center}
\footnotesize
\setlength{\tabcolsep}{1.0pt}
\resizebox{\linewidth}{!}{
\begin{tabular}{P{0.08\linewidth}P{0.11\linewidth}p{0.11\linewidth}P{0.12\linewidth}P{0.12\linewidth}P{0.12\linewidth}P{0.12\linewidth}P{0.12\linewidth}P{0.12\linewidth}}
    \toprule
    \multirow{2}{*}[-5pt]{Metric} &\multirow{2}{*}[-5pt]{Model} &\multirow{2}{*}[-5pt]{Method} &\multicolumn{6}{c}{Dataset}\\
    \cmidrule{4-9}
    & & &CIFAR-100 &\specialcell[c]{Tiny\\ImageNet} &\specialcell[c]{CUB\\200-2011} &\specialcell[c]{Stanford\\Dogs} &MIT67 &\specialcell[c]{FGVC\\Aircraft}\\
    \midrule
    \midrule
    \multirow{12}{*}[-2pt]{\shortstack{Top-5\\Error}} &\multirow{7}{*}[-2pt]{\shortstack{ResNet\\18}} &Baseline 
    &6.80\fontsize{0.2cm}{0.2cm}\selectfont$\pm$0.02 &22.61\fontsize{0.2cm}{0.2cm}\selectfont$\pm$0.20 &15.33\fontsize{0.2cm}{0.2cm}\selectfont$\pm$0.27 &9.67\fontsize{0.2cm}{0.2cm}\selectfont$\pm$0.53 &16.19\fontsize{0.2cm}{0.2cm}\selectfont$\pm$0.27 &4.94\fontsize{0.2cm}{0.2cm}\selectfont$\pm$0.23\\
    & &CS-KD~\cite{Yun_2020_CVPR}
    &5.63\fontsize{0.2cm}{0.2cm}\selectfont$\pm$0.15 &\textbf{19.42\fontsize{0.2cm}{0.2cm}\selectfont$\pm$0.12} &13.74\fontsize{0.2cm}{0.2cm}\selectfont$\pm$0.05 &8.52\fontsize{0.2cm}{0.2cm}\selectfont$\pm$0.04 &16.97\fontsize{0.2cm}{0.2cm}\selectfont$\pm$0.44 &4.34\fontsize{0.2cm}{0.2cm}\selectfont$\pm$0.26\\
    & &TF-KD~\cite{yuan2020revisiting}
    &5.61\fontsize{0.2cm}{0.2cm}\selectfont$\pm$0.07 &21.17\fontsize{0.2cm}{0.2cm}\selectfont$\pm$0.30 &14.09\fontsize{0.2cm}{0.2cm}\selectfont$\pm$0.25 &8.78\fontsize{0.2cm}{0.2cm}\selectfont$\pm$0.32 &15.57\fontsize{0.2cm}{0.2cm}\selectfont$\pm$0.56 &4.74\fontsize{0.2cm}{0.2cm}\selectfont$\pm$0.13\\
    & &PS-KD~\cite{Kim_2021_ICCV}
    &\underline{5.29\fontsize{0.2cm}{0.2cm}\selectfont$\pm$0.12} &20.53\fontsize{0.2cm}{0.2cm}\selectfont$\pm$0.22 &\underline{13.67\fontsize{0.2cm}{0.2cm}\selectfont$\pm$0.53} &\underline{7.90\fontsize{0.2cm}{0.2cm}\selectfont$\pm$0.19} &\textbf{14.35\fontsize{0.2cm}{0.2cm}\selectfont$\pm$0.86} &\textbf{3.89\fontsize{0.2cm}{0.2cm}\selectfont$\pm$0.14}\\
    & &TF-FD~\cite{li2022self}
    &6.45\fontsize{0.2cm}{0.2cm}\selectfont$\pm$0.40 &22.67\fontsize{0.2cm}{0.2cm}\selectfont$\pm$0.25 &13.45\fontsize{0.2cm}{0.2cm}\selectfont$\pm$0.62 &9.91\fontsize{0.2cm}{0.2cm}\selectfont$\pm$0.09 &14.89\fontsize{0.2cm}{0.2cm}\selectfont$\pm$0.16 &4.38\fontsize{0.2cm}{0.2cm}\selectfont$\pm$0.21\\
    & &ZipfsLS~\cite{liang2022efficient}
    &6.88\fontsize{0.2cm}{0.2cm}\selectfont$\pm$0.28 &22.41\fontsize{0.2cm}{0.2cm}\selectfont$\pm$0.25 &16.12\fontsize{0.2cm}{0.2cm}\selectfont$\pm$0.97 &10.17\fontsize{0.2cm}{0.2cm}\selectfont$\pm$0.87 &18.22\fontsize{0.2cm}{0.2cm}\selectfont$\pm$0.26 &5.72\fontsize{0.2cm}{0.2cm}\selectfont$\pm$0.35\\
    \cmidrule{3-9}
    & &\textbf{AI-KD}
    &\textbf{4.81\fontsize{0.2cm}{0.2cm}\selectfont$\pm$0.04} &\underline{19.87\fontsize{0.2cm}{0.2cm}\selectfont$\pm$0.21} &\textbf{11.47\fontsize{0.2cm}{0.2cm}\selectfont$\pm$0.23} &\textbf{8.17\fontsize{0.2cm}{0.2cm}\selectfont$\pm$0.06} &\underline{15.55\fontsize{0.2cm}{0.2cm}\selectfont$\pm$0.37} &\underline{4.06\fontsize{0.2cm}{0.2cm}\selectfont$\pm$0.15}\\
    \cmidrule{2-9}
    &\multirow{7}{*}[-2pt]{\shortstack{ResNet\\50}} &Baseline 
    &5.55\fontsize{0.2cm}{0.2cm}\selectfont$\pm$0.26 &19.66\fontsize{0.2cm}{0.2cm}\selectfont$\pm$0.07 
    &14.75\fontsize{0.2cm}{0.2cm}\selectfont$\pm$0.09 &8.77\fontsize{0.2cm}{0.2cm}\selectfont$\pm$0.53 
    &13.51\fontsize{0.2cm}{0.2cm}\selectfont$\pm$0.84 &3.99\fontsize{0.2cm}{0.2cm}\selectfont$\pm$0.46\\
    & &CS-KD~\cite{Yun_2020_CVPR} 
    &5.79\fontsize{0.2cm}{0.2cm}\selectfont$\pm$0.19 &\underline{17.96\fontsize{0.2cm}{0.2cm}\selectfont$\pm$0.51} &16.94\fontsize{0.2cm}{0.2cm}\selectfont$\pm$0.23 &10.76\fontsize{0.2cm}{0.2cm}\selectfont$\pm$0.64 &19.86\fontsize{0.2cm}{0.2cm}\selectfont$\pm$0.74 &5.85\fontsize{0.2cm}{0.2cm}\selectfont$\pm$0.36\\
    & &TF-KD~\cite{yuan2020revisiting} 
    &\textbf{4.31\fontsize{0.2cm}{0.2cm}\selectfont$\pm$0.19} &18.64\fontsize{0.2cm}{0.2cm}\selectfont$\pm$0.27 &\underline{13.90\fontsize{0.2cm}{0.2cm}\selectfont$\pm$1.52} &\underline{6.07\fontsize{0.2cm}{0.2cm}\selectfont$\pm$0.26} &\textbf{11.87\fontsize{0.2cm}{0.2cm}\selectfont$\pm$0.63} &\underline{3.35\fontsize{0.2cm}{0.2cm}\selectfont$\pm$0.25}\\
    & &PS-KD~\cite{Kim_2021_ICCV}
    &4.73\fontsize{0.2cm}{0.2cm}\selectfont$\pm$0.25 &18.05\fontsize{0.2cm}{0.2cm}\selectfont$\pm$0.17 
    &16.96\fontsize{0.2cm}{0.2cm}\selectfont$\pm$0.13 &7.33\fontsize{0.2cm}{0.2cm}\selectfont$\pm$0.58 
    &15.01\fontsize{0.2cm}{0.2cm}\selectfont$\pm$0.74 &3.97\fontsize{0.2cm}{0.2cm}\selectfont$\pm$0.44\\
    & &TF-FD~\cite{li2022self} 
    &5.25\fontsize{0.2cm}{0.2cm}\selectfont$\pm$0.26 &18.02\fontsize{0.2cm}{0.2cm}\selectfont$\pm$0.10
    &15.66\fontsize{0.2cm}{0.2cm}\selectfont$\pm$0.12 &8.22\fontsize{0.2cm}{0.2cm}\selectfont$\pm$0.28 
    &14.63\fontsize{0.2cm}{0.2cm}\selectfont$\pm$0.73 &4.31\fontsize{0.2cm}{0.2cm}\selectfont$\pm$0.14\\
    & &ZipfsLS~\cite{liang2022efficient} 
    &6.77\fontsize{0.2cm}{0.2cm}\selectfont$\pm$0.05 &20.65\fontsize{0.2cm}{0.2cm}\selectfont$\pm$0.07 
    &18.38\fontsize{0.2cm}{0.2cm}\selectfont$\pm$0.06 &12.68\fontsize{0.2cm}{0.2cm}\selectfont$\pm$1.21 
    &21.27\fontsize{0.2cm}{0.2cm}\selectfont$\pm$1.12 &8.04\fontsize{0.2cm}{0.2cm}\selectfont$\pm$1.30\\
    \cmidrule{3-9}
    & &\textbf{AI-KD} 
    &\underline{4.58\fontsize{0.2cm}{0.2cm}\selectfont$\pm$0.33} &\textbf{17.95\fontsize{0.2cm}{0.2cm}\selectfont$\pm$0.38} &\textbf{12.14\fontsize{0.2cm}{0.2cm}\selectfont$\pm$0.37} &\textbf{6.58\fontsize{0.2cm}{0.2cm}\selectfont$\pm$0.09} &\underline{13.06\fontsize{0.2cm}{0.2cm}\selectfont$\pm$0.52} &\textbf{3.20\fontsize{0.2cm}{0.2cm}\selectfont$\pm$0.29}\\
    \midrule
    \multirow{12}{*}{ECE} &\multirow{8}{*}[-2pt]{\shortstack{ResNet\\18}} &Baseline &11.90 &\underline{10.51} &\textbf{7.18} &\textbf{7.64} &\textbf{12.80} &\textbf{3.98}\\
    & &CS-KD~\cite{Yun_2020_CVPR} &\underline{5.50} &\textbf{3.73} &15.45 &9.75 &15.58 &6.24\\
    & &TF-KD~\cite{yuan2020revisiting} &11.03 &10.64 &\underline{8.01} &\underline{8.13} &\underline{12.82} &\underline{4.97}\\
    & &PS-KD~\cite{Kim_2021_ICCV} &\textbf{1.37} &21.41 &24.88 &30.05 &20.44 &24.45\\
    & &TF-FD~\cite{li2022self} &11.41 &10.89 &5.45 &7.32 &10.31 &3.30\\
    & &ZipfsLS~\cite{liang2022efficient} &21.04 &13.57 &13.16 &5.77 &8.81 &8.55\\
    \cmidrule{3-9}
    & &\textbf{AI-KD} &14.10 &15.00 &26.43 &20.41 &20.29 &18.74\\
    & &{~+ TS~\cite{guo2017calibration}} &{2.95} &{1.63} &{2.80} &{3.39} &{4.80} &{1.89}\\    
    \cmidrule{2-9}
    &\multirow{7}{*}[-2pt]{\shortstack{ResNet\\50}} &Baseline 
    &9.74\fontsize{0.2cm}{0.2cm}\selectfont &9.15\fontsize{0.2cm}{0.2cm}\selectfont 
    &5.93\fontsize{0.2cm}{0.2cm}\selectfont &7.58\fontsize{0.2cm}{0.2cm}\selectfont 
    &8.92\fontsize{0.2cm}{0.2cm}\selectfont &2.28\fontsize{0.2cm}{0.2cm}\selectfont\\
    & &CS-KD~\cite{Yun_2020_CVPR} 
    &9.53\fontsize{0.2cm}{0.2cm}\selectfont &2.96\fontsize{0.2cm}{0.2cm}\selectfont
    &9.75\fontsize{0.2cm}{0.2cm}\selectfont &8.30\fontsize{0.2cm}{0.2cm}\selectfont
    &10.73\fontsize{0.2cm}{0.2cm}\selectfont &8.60\fontsize{0.2cm}{0.2cm}\selectfont\\
    & &TF-KD~\cite{yuan2020revisiting} 
    &15.92\fontsize{0.2cm}{0.2cm}\selectfont &11.27\fontsize{0.2cm}{0.2cm}\selectfont &9.75\fontsize{0.2cm}{0.2cm}\selectfont &9.66\fontsize{0.2cm}{0.2cm}\selectfont
    &7.93\fontsize{0.2cm}{0.2cm}\selectfont &2.41\fontsize{0.2cm}{0.2cm}\selectfont\\
    & &PS-KD~\cite{Kim_2021_ICCV}
    &4.83\fontsize{0.2cm}{0.2cm}\selectfont &26.41\fontsize{0.2cm}{0.2cm}\selectfont 
    &21.39\fontsize{0.2cm}{0.2cm}\selectfont &22.34\fontsize{0.2cm}{0.2cm}\selectfont 
    &14.03\fontsize{0.2cm}{0.2cm}\selectfont &26.69\fontsize{0.2cm}{0.2cm}\selectfont\\
    & &TF-FD~\cite{li2022self} 
    &9.72\fontsize{0.2cm}{0.2cm}\selectfont &8.87\fontsize{0.2cm}{0.2cm}\selectfont
    &7.69\fontsize{0.2cm}{0.2cm}\selectfont &6.34\fontsize{0.2cm}{0.2cm}\selectfont 
    &8.97\fontsize{0.2cm}{0.2cm}\selectfont &1.65\fontsize{0.2cm}{0.2cm}\selectfont\\
    & &ZipfsLS~\cite{liang2022efficient} 
    &16.24\fontsize{0.2cm}{0.2cm}\selectfont &12.60\fontsize{0.2cm}{0.2cm}\selectfont 
    &10.01\fontsize{0.2cm}{0.2cm}\selectfont &5.80\fontsize{0.2cm}{0.2cm}\selectfont 
    &6.65\fontsize{0.2cm}{0.2cm}\selectfont &12.15\fontsize{0.2cm}{0.2cm}\selectfont\\
    \cmidrule{3-9}
    & &\textbf{AI-KD} 
    &\textbf{11.86\fontsize{0.2cm}{0.2cm}\selectfont} &\textbf{17.31\fontsize{0.2cm}{0.2cm}\selectfont} &\textbf{22.48\fontsize{0.2cm}{0.2cm}\selectfont} &\textbf{18.39\fontsize{0.2cm}{0.2cm}\selectfont} &\textbf{17.04\fontsize{0.2cm}{0.2cm}\selectfont} &\textbf{17.75\fontsize{0.2cm}{0.2cm}\selectfont}\\
    \bottomrule
\end{tabular}}
\end{center}
\vspace*{-3mm}
\caption{The performance using Top-5 error rates, and expected calibration error (ECE)~\cite{guo2017calibration} on ResNet-18 and ResNet-50 with various datasets. We report mean and standard deviation over three runs. The best and second-best results are indicated in bold and underlined, respectively.
TS denotes the temperature scaling for calibration.}
\vspace*{-3mm}
\label{tab:calib}
\end{table*}
%-------------------------------------------------------------------------
\begin{table*}[!ht]
\begin{center}
\footnotesize
\setlength{\tabcolsep}{1.0pt}
\resizebox{\linewidth}{!}{
\begin{tabular}{P{0.08\linewidth}P{0.11\linewidth}p{0.11\linewidth}P{0.12\linewidth}P{0.12\linewidth}P{0.12\linewidth}P{0.12\linewidth}P{0.12\linewidth}P{0.12\linewidth}}
    \toprule
    \multirow{2}{*}[-5pt]{Model} &\multirow{2}{*}[-5pt]{Metric} &\multirow{2}{*}[-5pt]{Method} &\multicolumn{6}{c}{Dataset}\\
    \cmidrule{4-9}
    & & &CIFAR-100 &\specialcell[c]{Tiny\\ImageNet} &\specialcell[c]{CUB\\200-2011} &\specialcell[c]{Stanford\\Dogs} &MIT67 &\specialcell[c]{FGVC\\Aircraft}\\
    \midrule
    \midrule
    \multirow{20}{*}[-2pt]{\shortstack{DenseNet\\121}}&\multirow{5}{*}[-2pt]{\shortstack{Top-1\\Error}} &Baseline &19.71\fontsize{0.2cm}{0.2cm}\selectfont$\pm$0.37 &40.38\fontsize{0.2cm}{0.2cm}\selectfont$\pm$0.14 &34.69\fontsize{0.2cm}{0.2cm}\selectfont$\pm$0.09     &33.19\fontsize{0.2cm}{0.2cm}\selectfont$\pm$0.80     &37.59\fontsize{0.2cm}{0.2cm}\selectfont$\pm$0.62  &17.71\fontsize{0.2cm}{0.2cm}\selectfont$\pm$0.17\\
    & &CS-KD~\cite{Yun_2020_CVPR} &21.48\fontsize{0.2cm}{0.2cm}\selectfont$\pm$0.32 &\textbf{38.21\fontsize{0.2cm}{0.2cm}\selectfont$\pm$0.17} &\underline{30.93\fontsize{0.2cm}{0.2cm}\selectfont$\pm$0.82} &\underline{28.70\fontsize{0.2cm}{0.2cm}\selectfont$\pm$0.14} &41.02\fontsize{0.2cm}{0.2cm}\selectfont$\pm$1.28 &17.37\fontsize{0.2cm}{0.2cm}\selectfont$\pm$0.36\\
    & &TF-KD~\cite{yuan2020revisiting} &18.90\fontsize{0.2cm}{0.2cm}\selectfont$\pm$0.22 &38.46\fontsize{0.2cm}{0.2cm}\selectfont$\pm$0.36 &32.95\fontsize{0.2cm}{0.2cm}\selectfont$\pm$0.65 &30.41\fontsize{0.2cm}{0.2cm}\selectfont$\pm$0.29 &36.42\fontsize{0.2cm}{0.2cm}\selectfont$\pm$0.66 &16.71\fontsize{0.2cm}{0.2cm}\selectfont$\pm$0.40\\
    & &PS-KD~\cite{Kim_2021_ICCV} &\underline{18.69\fontsize{0.2cm}{0.2cm}\selectfont$\pm$0.12} &38.62\fontsize{0.2cm}{0.2cm}\selectfont$\pm$0.44 &34.28\fontsize{0.2cm}{0.2cm}\selectfont$\pm$0.50 &30.54\fontsize{0.2cm}{0.2cm}\selectfont$\pm$0.03 &\underline{35.92\fontsize{0.2cm}{0.2cm}\selectfont$\pm$0.48} &\underline{16.38\fontsize{0.2cm}{0.2cm}\selectfont$\pm$0.71}\\
    \cmidrule{3-9}
    & &\textbf{AI-KD} &\textbf{18.35\fontsize{0.2cm}{0.2cm}\selectfont$\pm$0.18} &\underline{38.33\fontsize{0.2cm}{0.2cm}\selectfont$\pm$0.07} &\textbf{26.65\fontsize{0.2cm}{0.2cm}\selectfont$\pm$0.29} &\textbf{27.13\fontsize{0.2cm}{0.2cm}\selectfont$\pm$0.16} &\textbf{34.58\fontsize{0.2cm}{0.2cm}\selectfont$\pm$0.75} &\textbf{13.84\fontsize{0.2cm}{0.2cm}\selectfont$\pm$0.38}\\    
    
    \cmidrule{2-9}
    &\multirow{5}{*}{\shortstack{F1 Score\\Macro}} &Baseline  &0.802\fontsize{0.2cm}{0.2cm}\selectfont$\pm$0.004 &0.594\fontsize{0.2cm}{0.2cm}\selectfont$\pm$0.001 &0.650\fontsize{0.2cm}{0.2cm}\selectfont$\pm$0.007 &0.656\fontsize{0.2cm}{0.2cm}\selectfont$\pm$0.007 &0.620\fontsize{0.2cm}{0.2cm}\selectfont$\pm$0.008 &0.824\fontsize{0.2cm}{0.2cm}\selectfont$\pm$0.002\\
    & &CS-KD~\cite{Yun_2020_CVPR} &0.784\fontsize{0.2cm}{0.2cm}\selectfont$\pm$0.005 &0.616\fontsize{0.2cm}{0.2cm}\selectfont$\pm$0.001 &\underline{0.689\fontsize{0.2cm}{0.2cm}\selectfont$\pm$0.007} &\underline{0.703\fontsize{0.2cm}{0.2cm}\selectfont$\pm$0.002} &0.588\fontsize{0.2cm}{0.2cm}\selectfont$\pm$0.014 &0.826\fontsize{0.2cm}{0.2cm}\selectfont$\pm$0.005\\
    & &TF-KD~\cite{yuan2020revisiting} &0.811\fontsize{0.2cm}{0.2cm}\selectfont$\pm$0.002 &\textbf{0.617\fontsize{0.2cm}{0.2cm}\selectfont$\pm$0.002} &0.668\fontsize{0.2cm}{0.2cm}\selectfont$\pm$0.007 &0.686\fontsize{0.2cm}{0.2cm}\selectfont$\pm$0.004 &0.633\fontsize{0.2cm}{0.2cm}\selectfont$\pm$0.006 &0.833\fontsize{0.2cm}{0.2cm}\selectfont$\pm$0.005\\
    & &PS-KD~\cite{Kim_2021_ICCV} &\underline{0.813\fontsize{0.2cm}{0.2cm}\selectfont$\pm$0.001} &0.609\fontsize{0.2cm}{0.2cm}\selectfont$\pm$0.005 &0.655\fontsize{0.2cm}{0.2cm}\selectfont$\pm$0.005 &0.681\fontsize{0.2cm}{0.2cm}\selectfont$\pm$0.001 &\underline{0.635\fontsize{0.2cm}{0.2cm}\selectfont$\pm$0.004} &\underline{0.835\fontsize{0.2cm}{0.2cm}\selectfont$\pm$0.008}\\
    \cmidrule{3-9}
    & &\textbf{AI-KD} &\textbf{0.816\fontsize{0.2cm}{0.2cm}\selectfont$\pm$0.002} &\textbf{0.617\fontsize{0.2cm}{0.2cm}\selectfont$\pm$0.002} &\textbf{0.730\fontsize{0.2cm}{0.2cm}\selectfont$\pm$0.004} &\textbf{0.714\fontsize{0.2cm}{0.2cm}\selectfont$\pm$0.006} &\textbf{0.649\fontsize{0.2cm}{0.2cm}\selectfont$\pm$0.008} &\textbf{0.861\fontsize{0.2cm}{0.2cm}\selectfont$\pm$0.004}\\
    \cmidrule{2-9}
    &\multirow{5}{*}[-2pt]{\shortstack{Top-5\\Error}} &Baseline &4.77\fontsize{0.2cm}{0.2cm}\selectfont$\pm$0.15 &17.73\fontsize{0.2cm}{0.2cm}\selectfont$\pm$0.19 &12.94\fontsize{0.2cm}{0.2cm}\selectfont$\pm$0.47 &8.56\fontsize{0.2cm}{0.2cm}\selectfont$\pm$0.50 &14.75\fontsize{0.2cm}{0.2cm}\selectfont$\pm$0.78 &4.31\fontsize{0.2cm}{0.2cm}\selectfont$\pm$0.18\\
    & &CS-KD~\cite{Yun_2020_CVPR} &6.67\fontsize{0.2cm}{0.2cm}\selectfont$\pm$0.08 &16.82\fontsize{0.2cm}{0.2cm}\selectfont$\pm$0.21 &\underline{11.36\fontsize{0.2cm}{0.2cm}\selectfont$\pm$0.38} &7.23\fontsize{0.2cm}{0.2cm}\selectfont$\pm$0.19 &16.52\fontsize{0.2cm}{0.2cm}\selectfont$\pm$0.80 &4.06\fontsize{0.2cm}{0.2cm}\selectfont$\pm$0.11\\
    & &TF-KD~\cite{yuan2020revisiting} &4.20\fontsize{0.2cm}{0.2cm}\selectfont$\pm$0.07 &\textbf{16.32\fontsize{0.2cm}{0.2cm}\selectfont$\pm$0.40} &11.98\fontsize{0.2cm}{0.2cm}\selectfont$\pm$0.11 &\underline{7.20\fontsize{0.2cm}{0.2cm}\selectfont$\pm$0.30} &\underline{12.61\fontsize{0.2cm}{0.2cm}\selectfont$\pm$0.27} &3.74\fontsize{0.2cm}{0.2cm}\selectfont$\pm$0.17\\
    & &PS-KD~\cite{Kim_2021_ICCV} &\textbf{4.06\fontsize{0.2cm}{0.2cm}\selectfont$\pm$0.11} &17.01\fontsize{0.2cm}{0.2cm}\selectfont$\pm$0.30 &13.18\fontsize{0.2cm}{0.2cm}\selectfont$\pm$0.34 &7.76\fontsize{0.2cm}{0.2cm}\selectfont$\pm$0.15 &13.18\fontsize{0.2cm}{0.2cm}\selectfont$\pm$0.34 &\underline{3.59\fontsize{0.2cm}{0.2cm}\selectfont$\pm$0.08}\\
    \cmidrule{3-9}    
    & &\textbf{AI-KD} &\underline{4.14\fontsize{0.2cm}{0.2cm}\selectfont$\pm$0.06} &\underline{16.50\fontsize{0.2cm}{0.2cm}\selectfont$\pm$0.44} &\textbf{9.46\fontsize{0.2cm}{0.2cm}\selectfont$\pm$0.14} &\textbf{6.34\fontsize{0.2cm}{0.2cm}\selectfont$\pm$0.13} &\textbf{12.44\fontsize{0.2cm}{0.2cm}\selectfont$\pm$0.82} &\textbf{3.53\fontsize{0.2cm}{0.2cm}\selectfont$\pm$0.09}\\
    \cmidrule{2-9}
    &\multirow{6}{*}{ECE} &Baseline &7.74 &4.60 &\underline{5.06} &4.69 &\underline{9.63} &\textbf{2.38}\\
    & &CS-KD~\cite{Yun_2020_CVPR} &14.66 &\underline{2.14} &7.27 &\textbf{2.83} &\textbf{8.74} &3.14\\
    & &TF-KD~\cite{yuan2020revisiting} &9.87 &5.81 &\textbf{4.18} &\underline{4.48} &10.64 &\underline{2.75}\\
    & &PS-KD~\cite{Kim_2021_ICCV} &\textbf{1.34} &24.16 &26.57 &33.64 &20.56 &27.80\\
    \cmidrule{3-9}
    & &\textbf{AI-KD} &\underline{5.11} &\textbf{1.89} &26.66 &20.14 &19.26 &18.20\\
    & &\textbf{~+ TS~\cite{guo2017calibration}} &1.70 &1.67 &1.99 &1.73 &4.06 &1.46\\    
    \bottomrule
\end{tabular}}
\end{center}
\vspace*{-3mm}
\caption{Image classification Top-1 error, F1 Score, Top-5 error results and expected calibration error (ECE) on DenseNet-121 with various datasets. We report mean and standard deviation over three runs. The best and second-best results are indicated in bold and underlined, respectively. TS denotes the
temperature scaling for calibration}
\vspace*{-3mm}
\label{tab:coarse_t1_dense}
\end{table*}
%-------------------------------------------------------------------------
\subsection{Comparison with representative Self-KD}
\label{sec:exp_classification}
As presented in Table~\ref{tab:coarse_t1}, AI-KD shows competitive performances on coarse datasets in two network architectures. Especially on the CIFAR-100 dataset, our proposed method records $19.87\%$ Top-1 error on PreAct ResNet-18 and outperforms CS-KD, TF-KD,
PS-KD, TF-FD, and ZipfsLS by $1.43\%$, $2.20\%$, $1.29\%$, $3.90\%$, and $2.30\%$, respectively.
Although our proposed method achieves the second-best Top-1 error rate on Preact ResNet-18 with Tiny ImageNet, it demonstrates improved performance compared to other methods when evaluated on Preact ResNet-50 with CIFAR-100 and Tiny ImageNet datasets. The observed trends in these results are also consistent with the F1 score macro metric.

Most of the remarkable points are evaluated performances of fine-grained datasets.
We experiment to evaluate AI-KD with four different fine-grained datasets on ResNet-18 and ResNet-50.
Our proposed method records outperforming performance in every result than other compared methods, while previous state-of-the-art methods such as CS-KD and PS-KD record the second-best performance, depending on specific datasets and network architectures. 
In particular, TF-KD demonstrates enhanced performance in ResNet-50 on the fine-grained dataset, attributed to its effective use of knowledge transferred from the ImageNet pre-trained model acting as the teacher than compared approaches. Nevertheless, AI-KD demonstrates outstanding performance on ResNet-18 and ResNet-50 with various fine-grained datasets, as evidenced by its low Top-1 error rate and high F1 score.

To further evaluate our proposed method, Table~\ref{tab:calib} shows the comparison results in terms of Top-5 error rate and ECE. AI-KD provides the best Top-5 error rate with CIFAR-100, CUB200-2011, and Stanford Dogs datasets on ResNet-18.
In ResNet-50, our method achieves the best results in terms of Top-5 error rate across the Tiny ImageNet, CUB200-2011, Stanford Dogs, and Aircraft datasets.
Additionally, we conducted experiments to assess the efficiency of our model using a different architecture, DenseNet-121, across a broader range of model configurations. As shown in Table~\ref{tab:coarse_t1_dense}, AI-KD records well-generalized results across all datasets, including CIFAR-100 and fine-grained datasets.
%-------------------------------------------------------------------------
\begin{figure*}[!t]
    \centering
    \resizebox{1.\linewidth}{!}{
    \setlength{\tabcolsep}{1pt}
    \begin{tabular}{ccc}
        \includegraphics[width=1\linewidth]{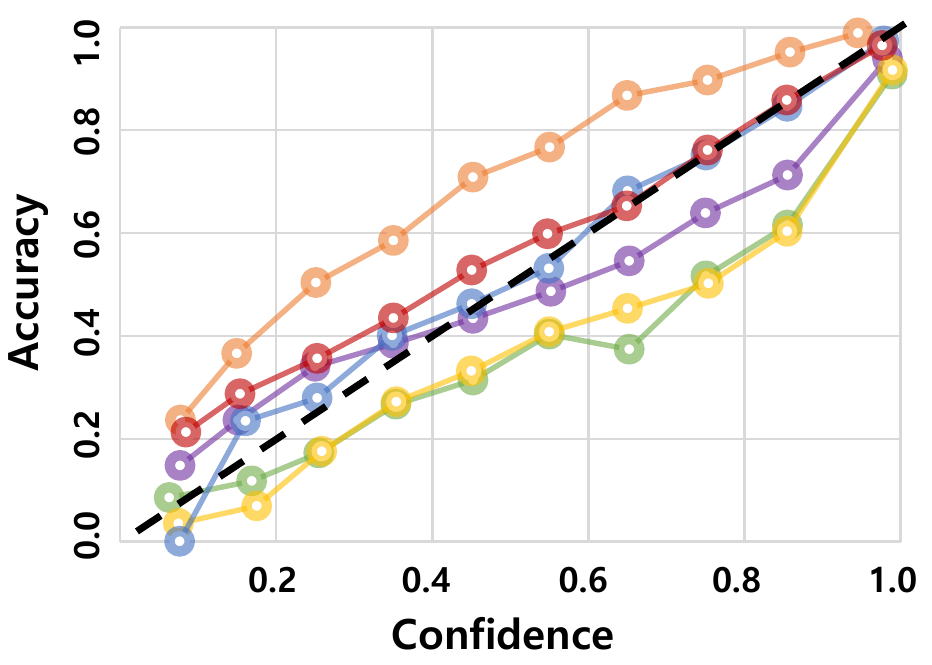} &\includegraphics[width=1\linewidth]{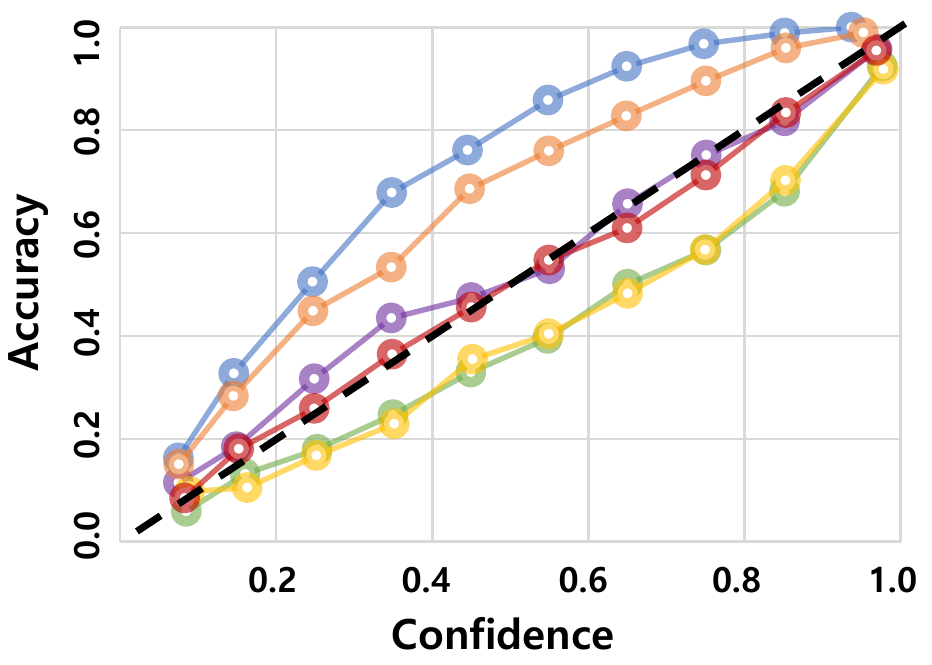} &\includegraphics[width=1\linewidth]{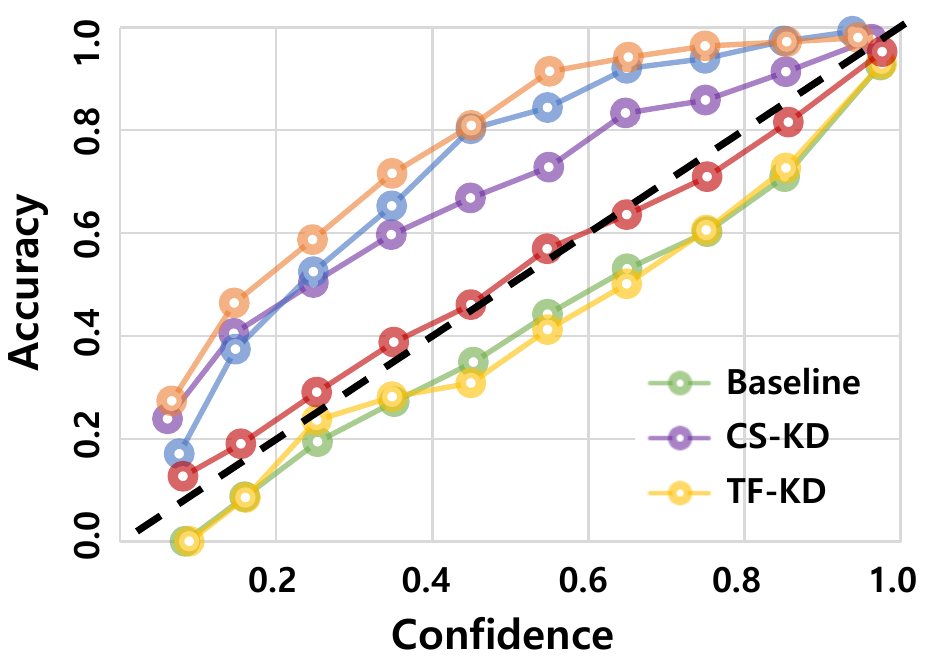}\\
        \fontsize{0.8cm}{0.8cm}\selectfont{(a) CIFAR-100} &\fontsize{0.8cm}{0.8cm}\selectfont{(b) Tiny ImageNet} &\fontsize{0.8cm}{0.8cm}\selectfont{(c) CUB200-2011}\\
        \includegraphics[width=1\linewidth]{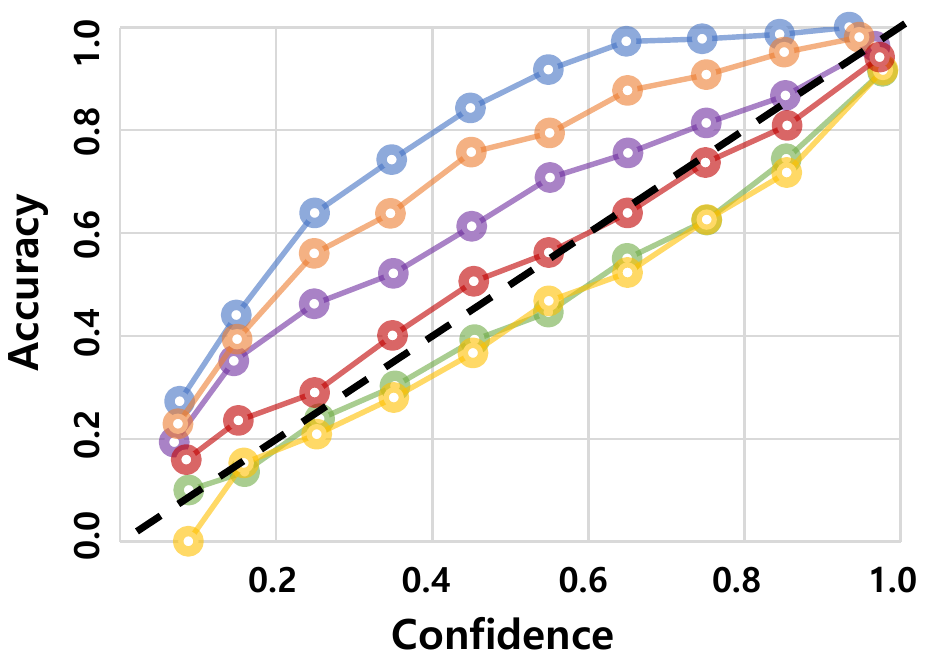}        &\includegraphics[width=1\linewidth]{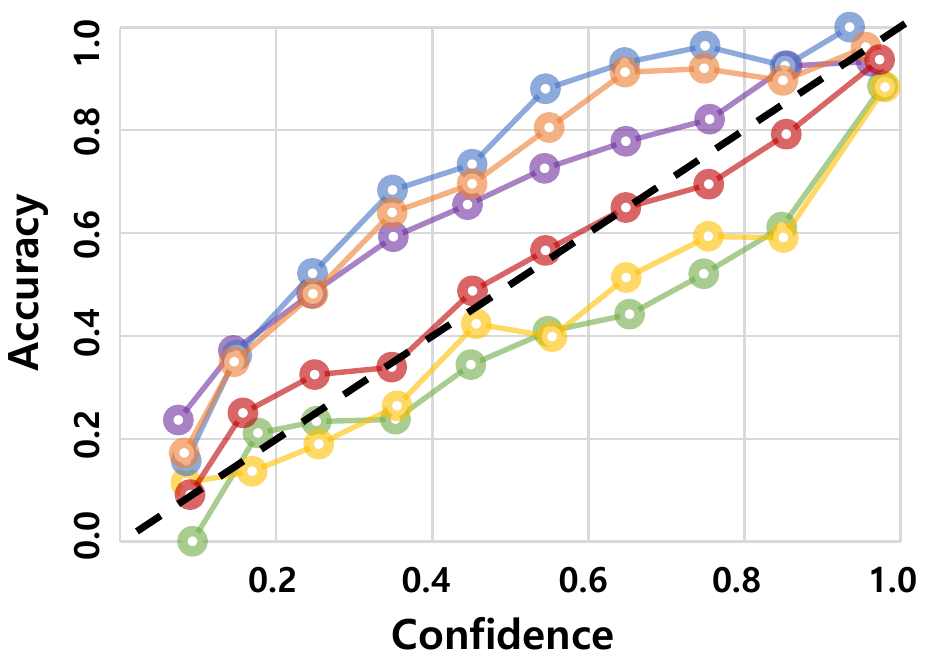}        &\includegraphics[width=1\linewidth]{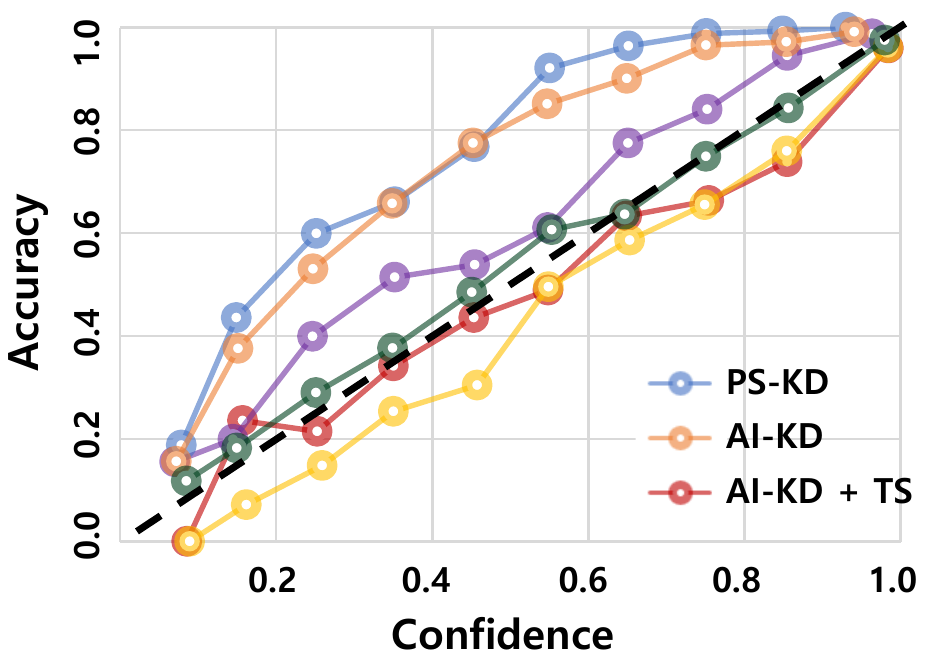}\\
        \fontsize{0.8cm}{0.8cm}\selectfont{(d) Stanford Dogs} &\fontsize{0.8cm}{0.8cm}\selectfont{(e) MIT67} &\fontsize{0.8cm}{0.8cm}\selectfont{(f) FGVC Aircraft}\\
    \end{tabular}}
    \caption{Confidence reliability diagrams based on ResNet-18 with various datasets.}
    \label{fig:calib}
\end{figure*}
%-------------------------------------------------------------------------
\begin{figure*}[!t]
    \centering
    \resizebox{1.\linewidth}{!}{
    \setlength{\tabcolsep}{1pt}
    \begin{tabular}{ccc}
        \includegraphics[width=1\linewidth]{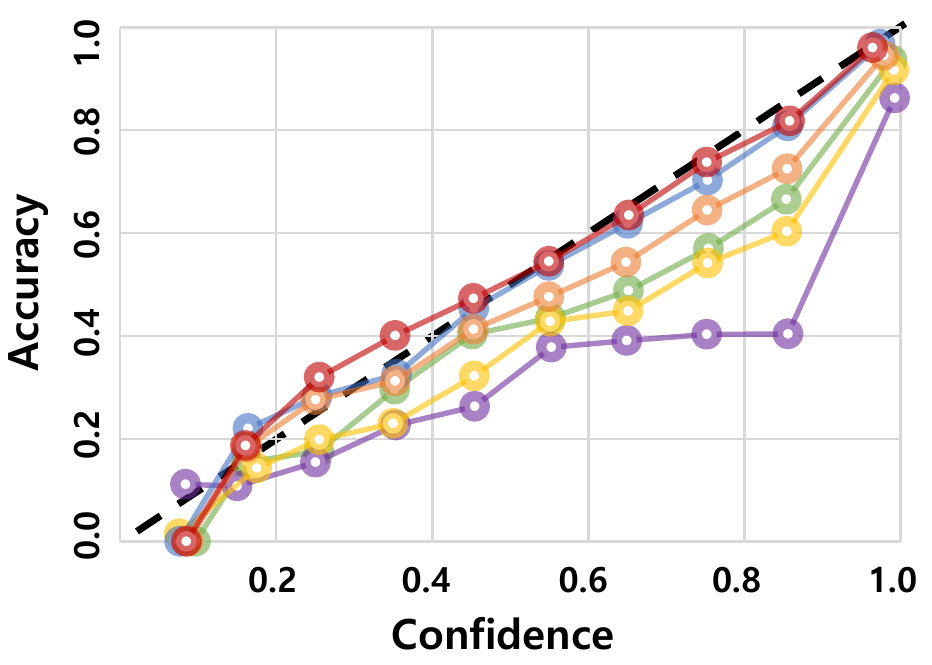} &\includegraphics[width=1\linewidth]{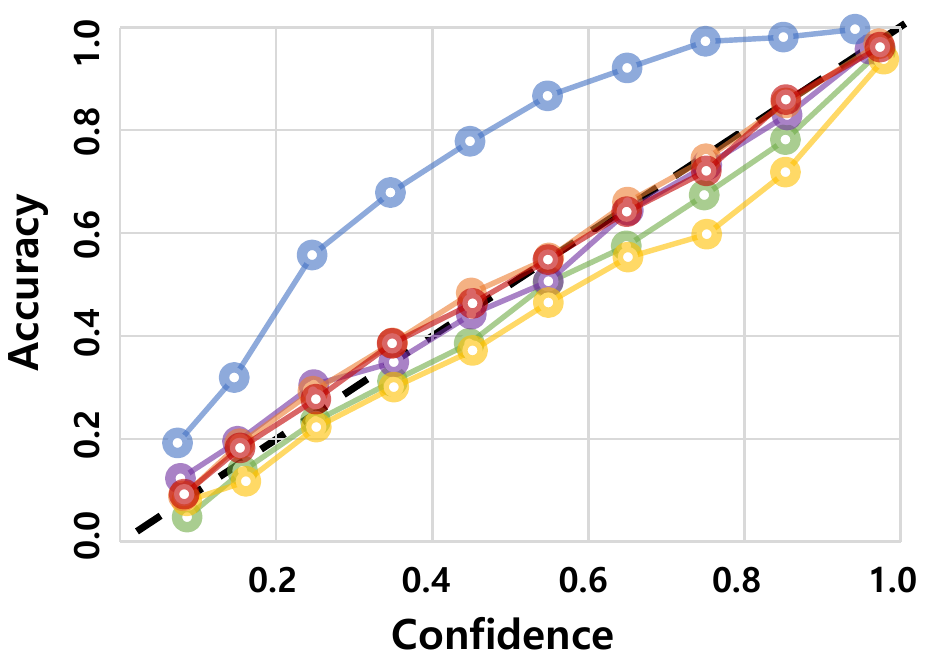} &\includegraphics[width=1\linewidth]{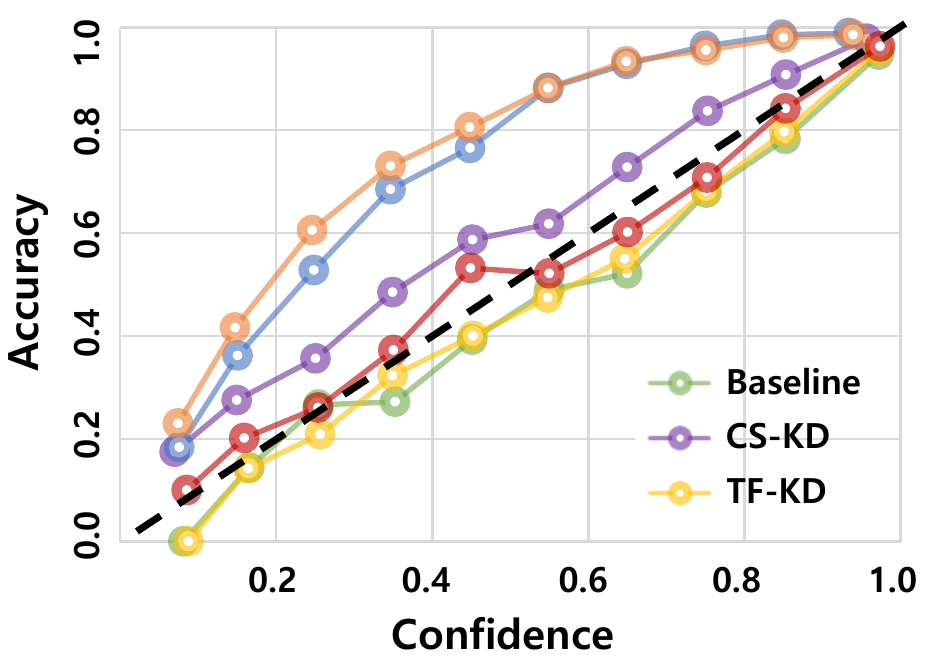}\\
        \fontsize{0.8cm}{0.8cm}\selectfont{(a) CIFAR-100} &\fontsize{0.8cm}{0.8cm}\selectfont{(b) Tiny ImageNet} &\fontsize{0.8cm}{0.8cm}\selectfont{(c) CUB200-2011}\\
        \includegraphics[width=1\linewidth]{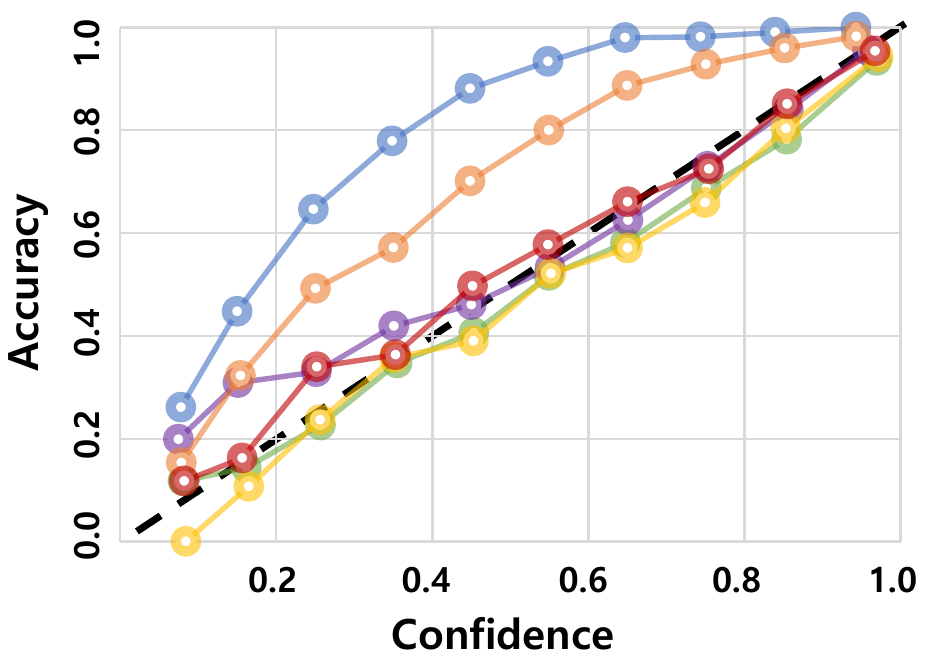}        &\includegraphics[width=1\linewidth]{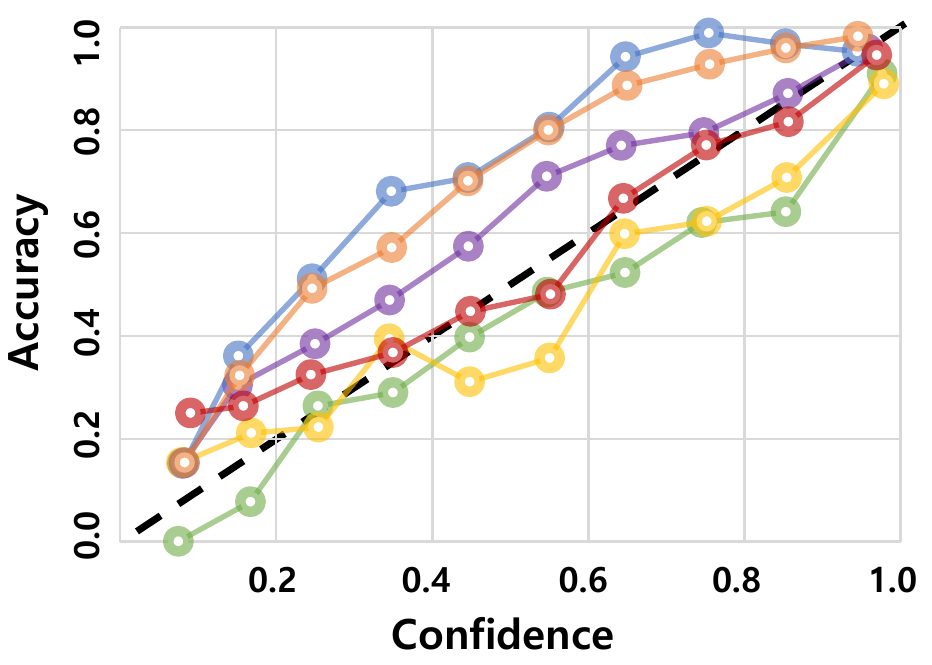}        &\includegraphics[width=1\linewidth]{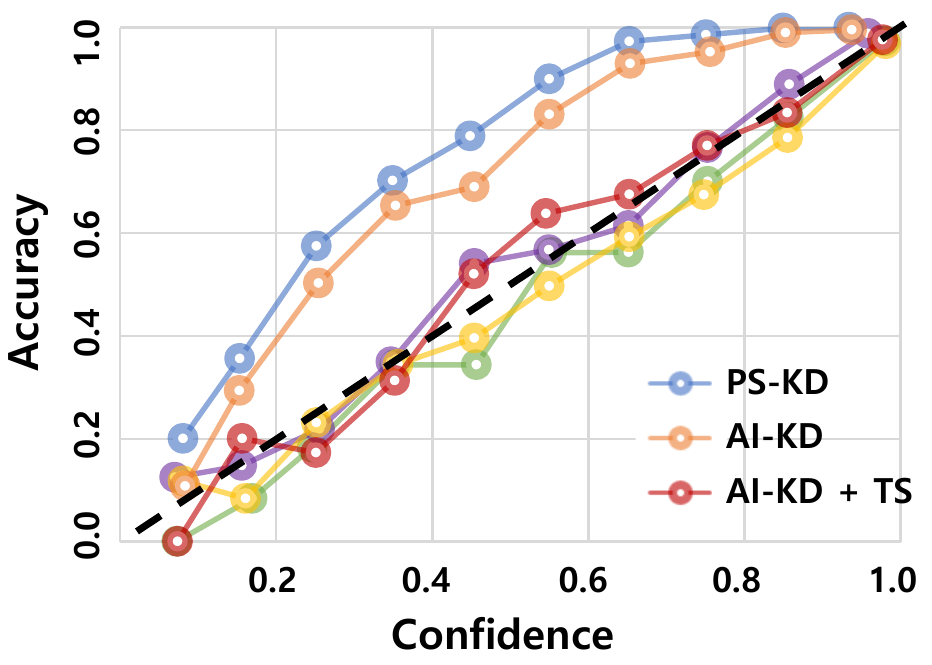}\\
        \fontsize{0.8cm}{0.8cm}\selectfont{(d) Stanford Dogs} &\fontsize{0.8cm}{0.8cm}\selectfont{(e) MIT67} &\fontsize{0.8cm}{0.8cm}\selectfont{(f) FGVC Aircraft}
    \end{tabular}}
    \caption{Confidence reliability diagrams based on DenseNet-121 with various datasets.}
    \label{fig:calib_b}
\end{figure*}

%-------------------------------------------------------------------------
Unlike other metrics, ECE is the metric to measure a model based on calibration error and is calculated as a weighted average over the absolute difference between accuracy and confidence. 
AI-KD shows tendencies to be over-confident under all the datasets and network models, except for an evaluation with Tiny-ImageNet on DenseNet-121.
However, the drawback can alleviate and prevent degradation using the calibration approach by temperature scaling~\cite{guo2017calibration}.
We conduct additional experiments to employ the calibration method and confirm the results. Applying the temperature scaling, AI-KD records well-calibrated results and alleviates over-confidence, as shown in Table~\ref{tab:calib} TS rows.
For better intuitive analysis, we describe the ECE result using confidence reliability diagrams with the datasets on ResNet-18 and DenseNet-121. The plotted diagonal dot lines in each figure indicate the identity function. The closer each solid line by the considered method to the diagonal dot lines, the method indicates well-calibrated. Figure~\ref{fig:calib} describes the results on ResNet-18 with all evaluated datasets. The solid orange line indicates confidence in our proposed method and is plotted over the diagonal line with all datasets. Employing temperature scaling to AI-KD, we confirm that the result, the solid red line, is plotted close to the diagonal dot line.
We also plot the reliability diagrams of our proposed method on DenseNet-121, as shown in Figure~\ref{fig:calib_b} and confirm the effectiveness as same with the results of ResNet-18 network model.

\begin{figure*}[!t]
    \centering
    \resizebox{1.\linewidth}{!}{
    \setlength{\tabcolsep}{1pt}
    \renewcommand{\arraystretch}{2}
    \begin{tabular}{ccccc}
        \includegraphics[width=0.8\linewidth]{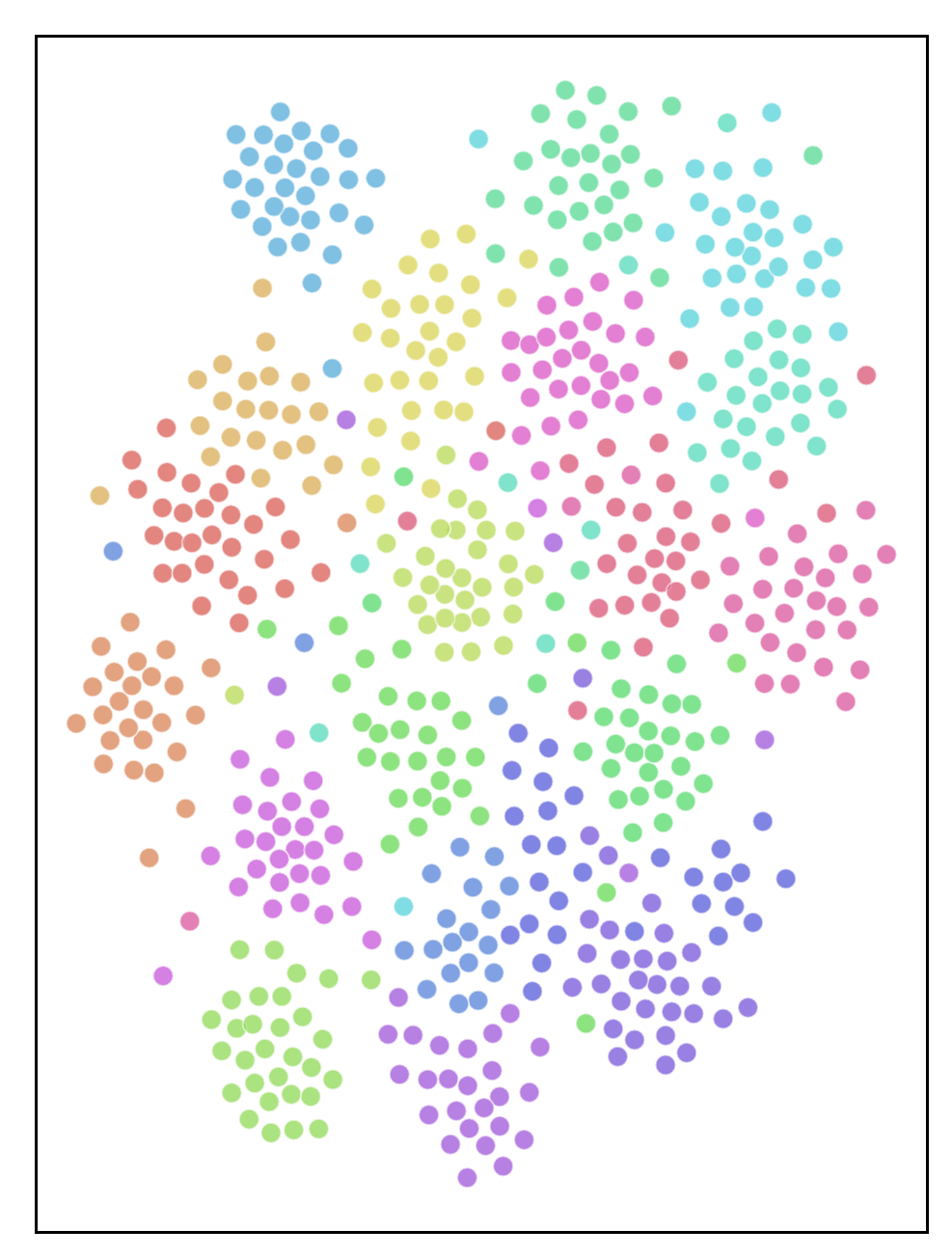} &\includegraphics[width=0.8\linewidth]{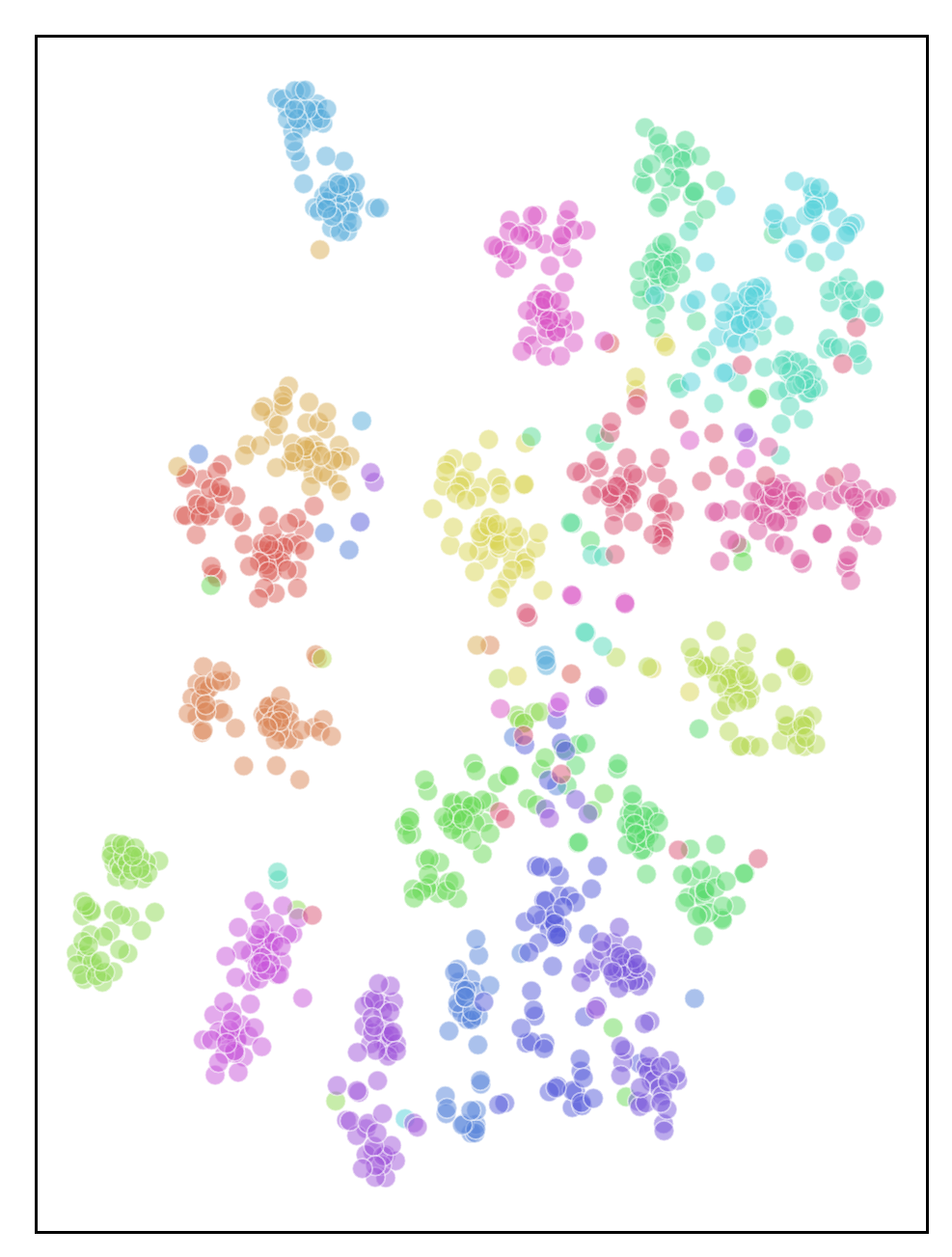} &\includegraphics[width=0.8\linewidth]{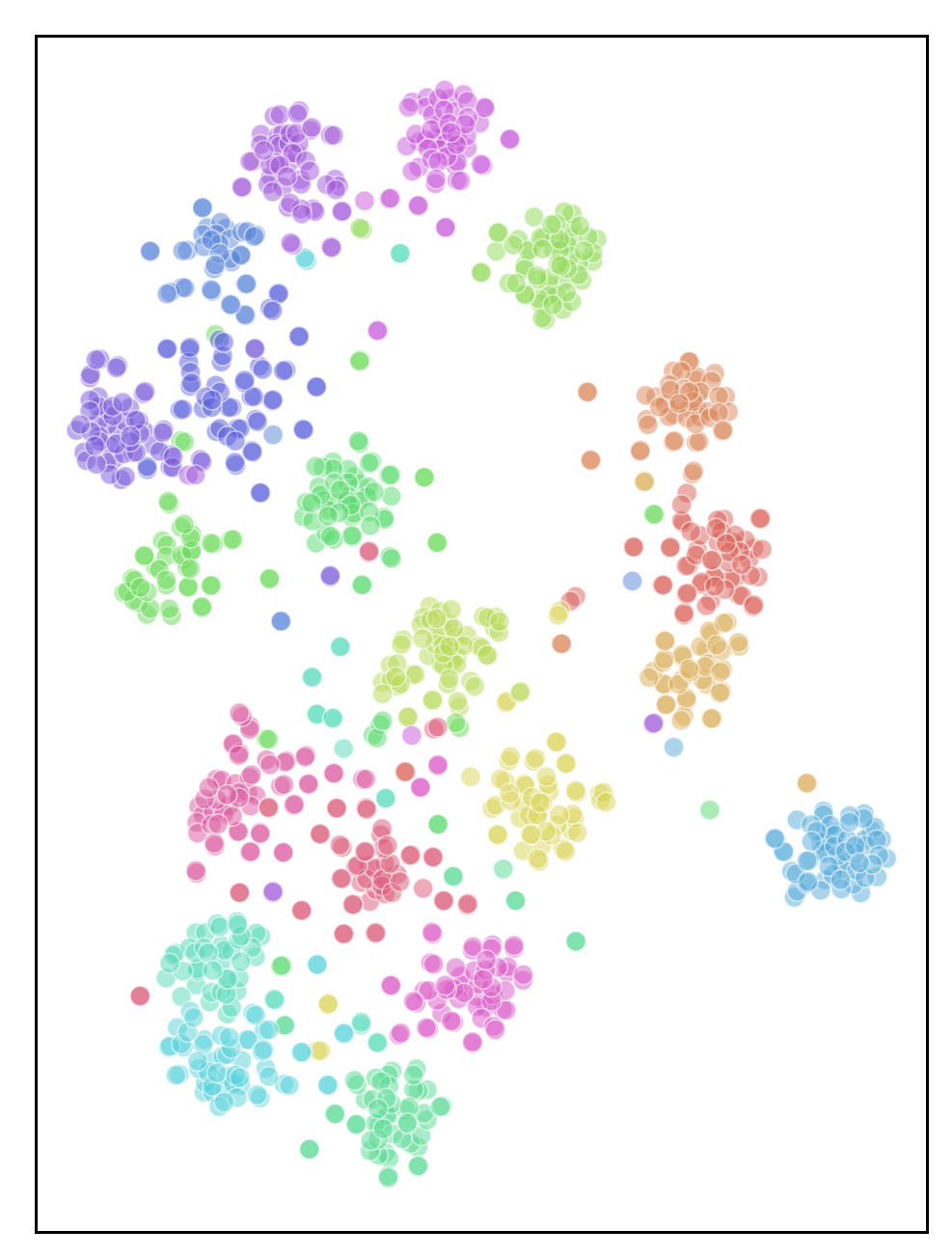} &\includegraphics[width=0.8\linewidth]{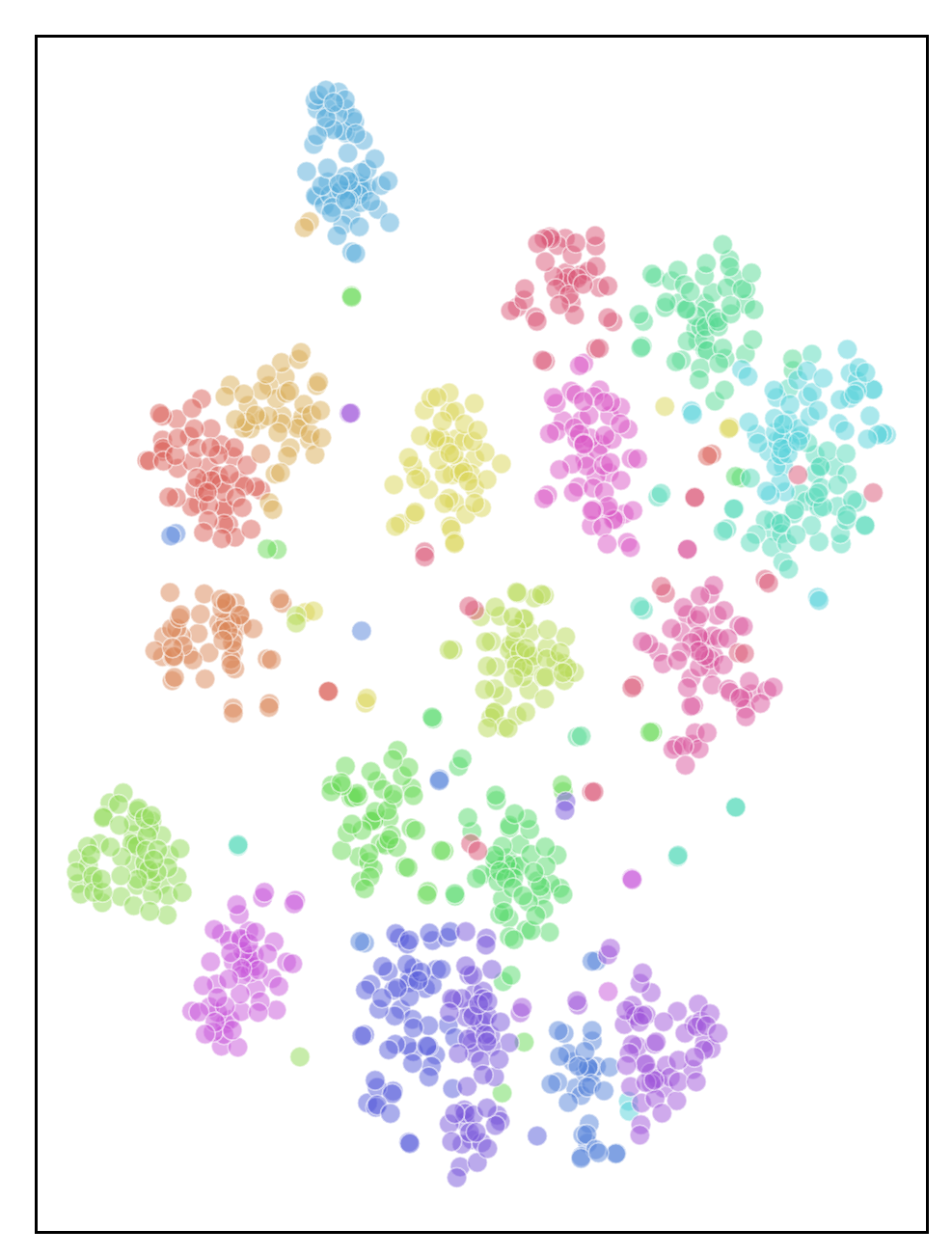} &\includegraphics[width=0.8\linewidth]{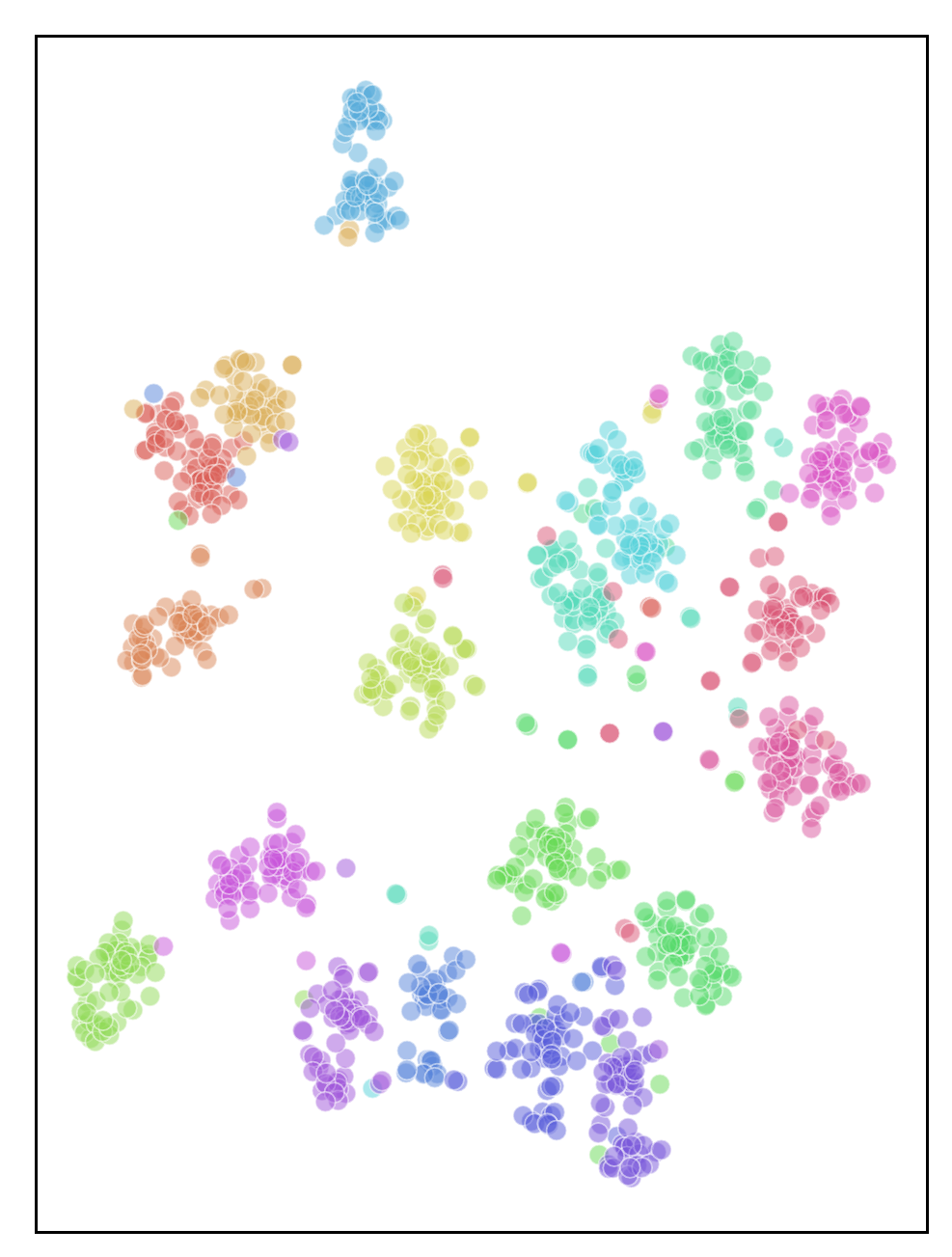}\\
        \includegraphics[width=0.8\linewidth]{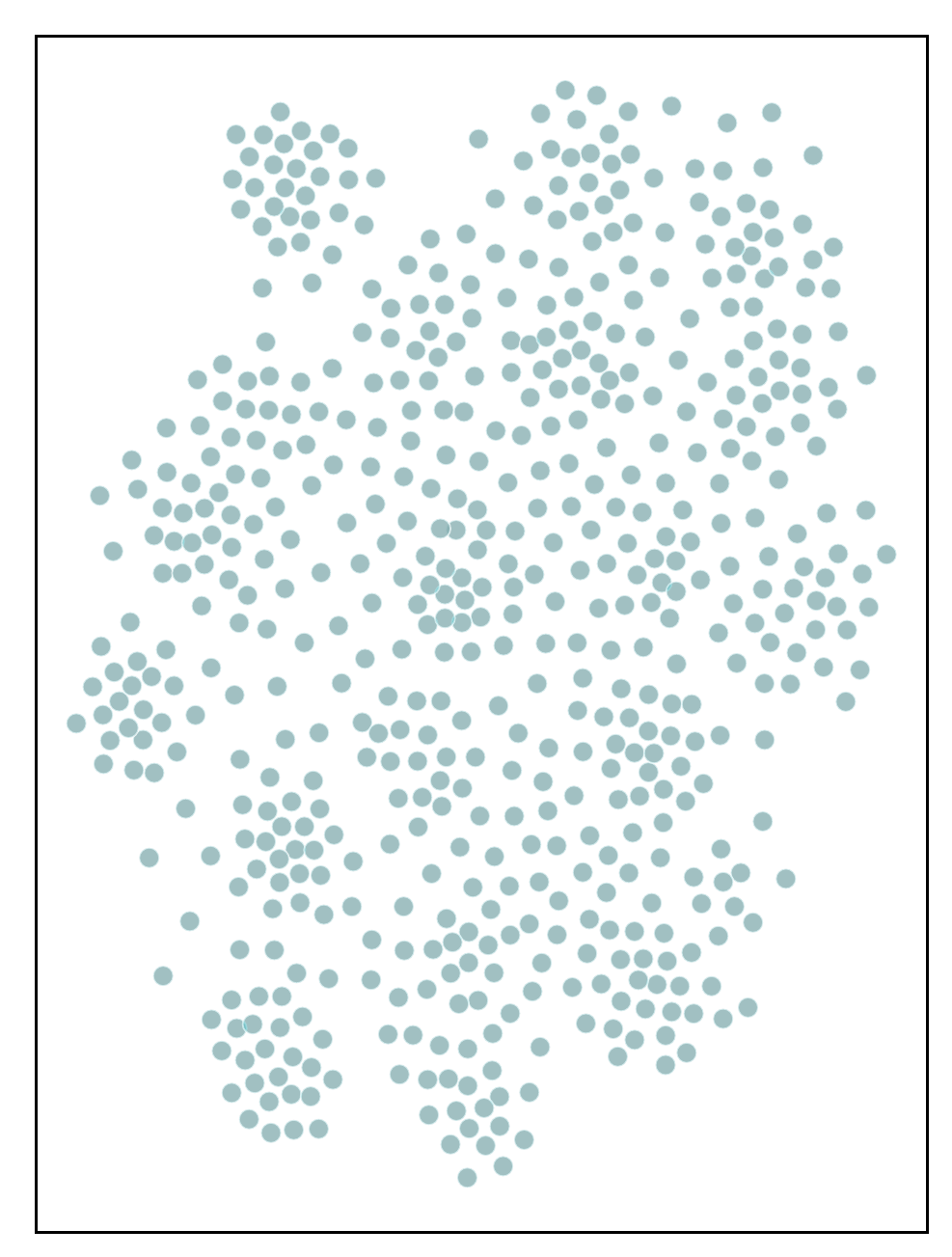} &\includegraphics[width=0.8\linewidth]{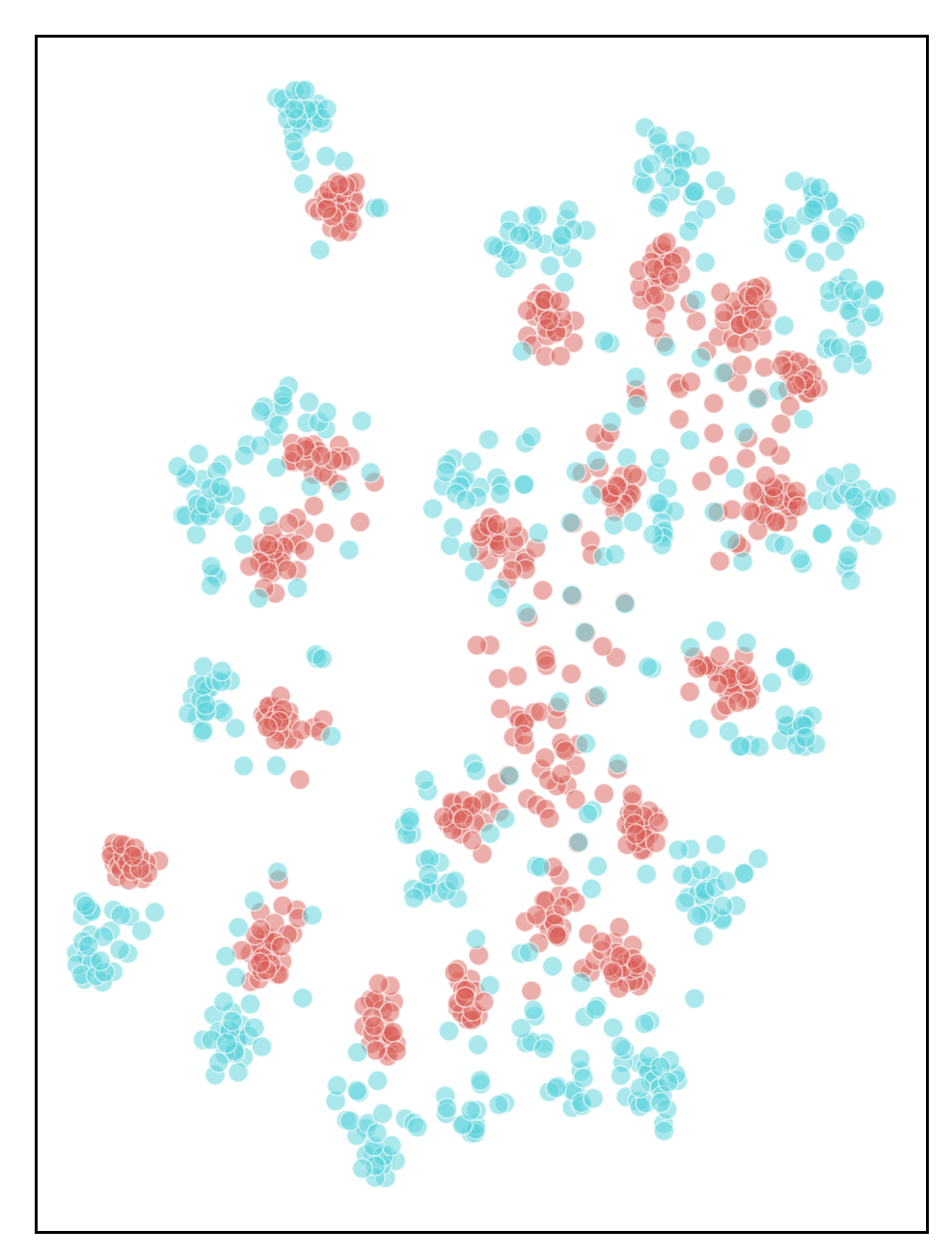} &\includegraphics[width=0.8\linewidth]{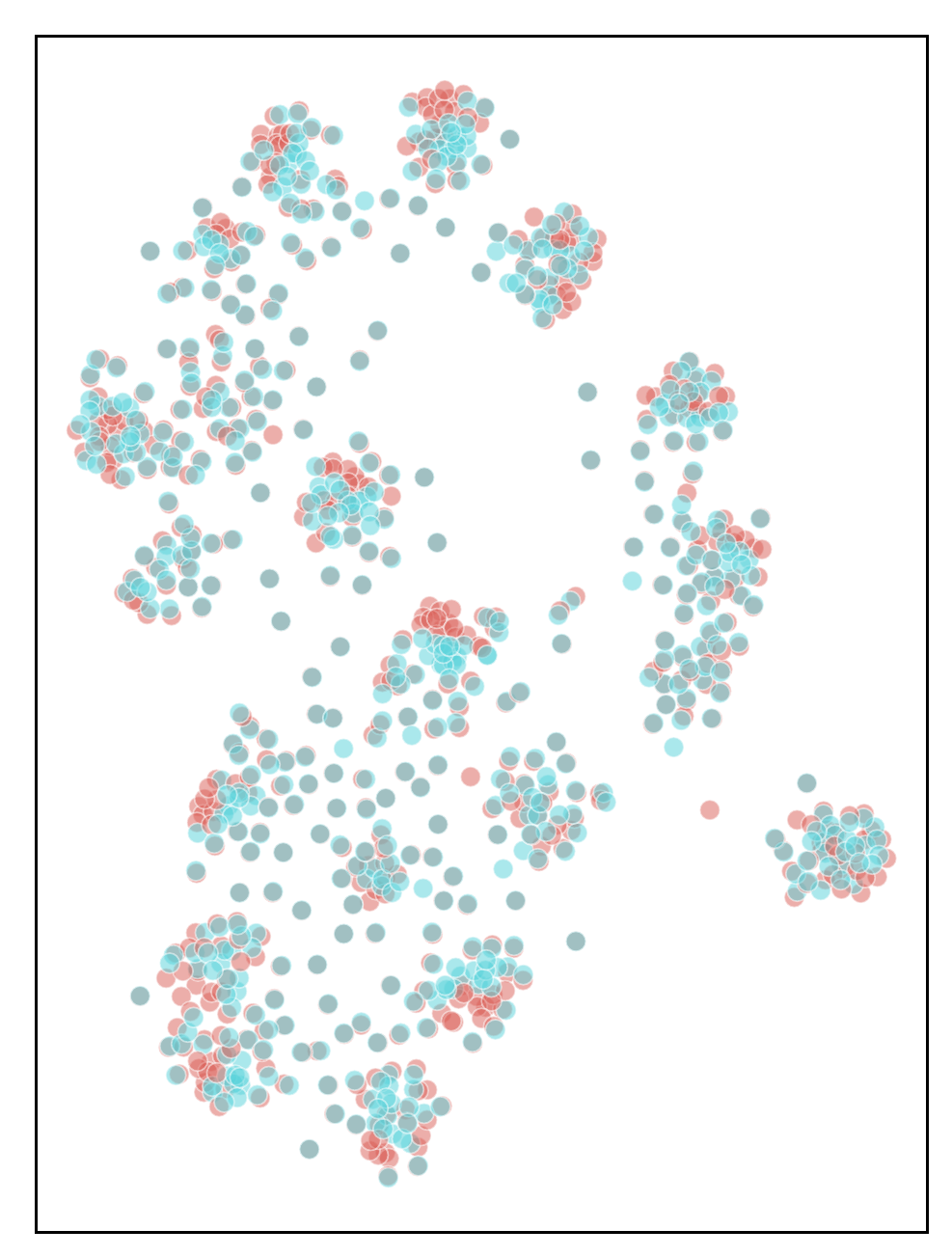} &\includegraphics[width=0.8\linewidth]{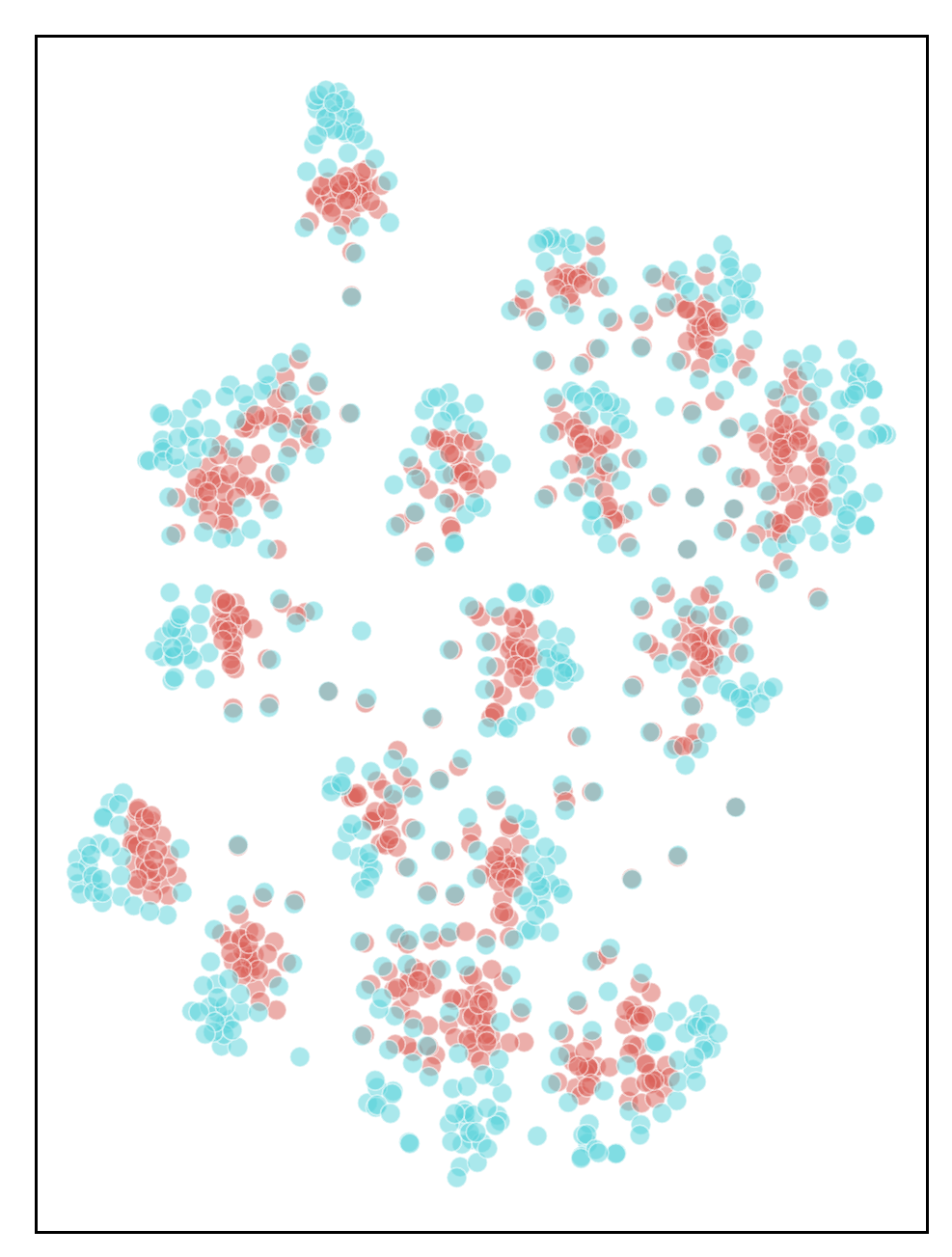} &\includegraphics[width=0.8\linewidth]{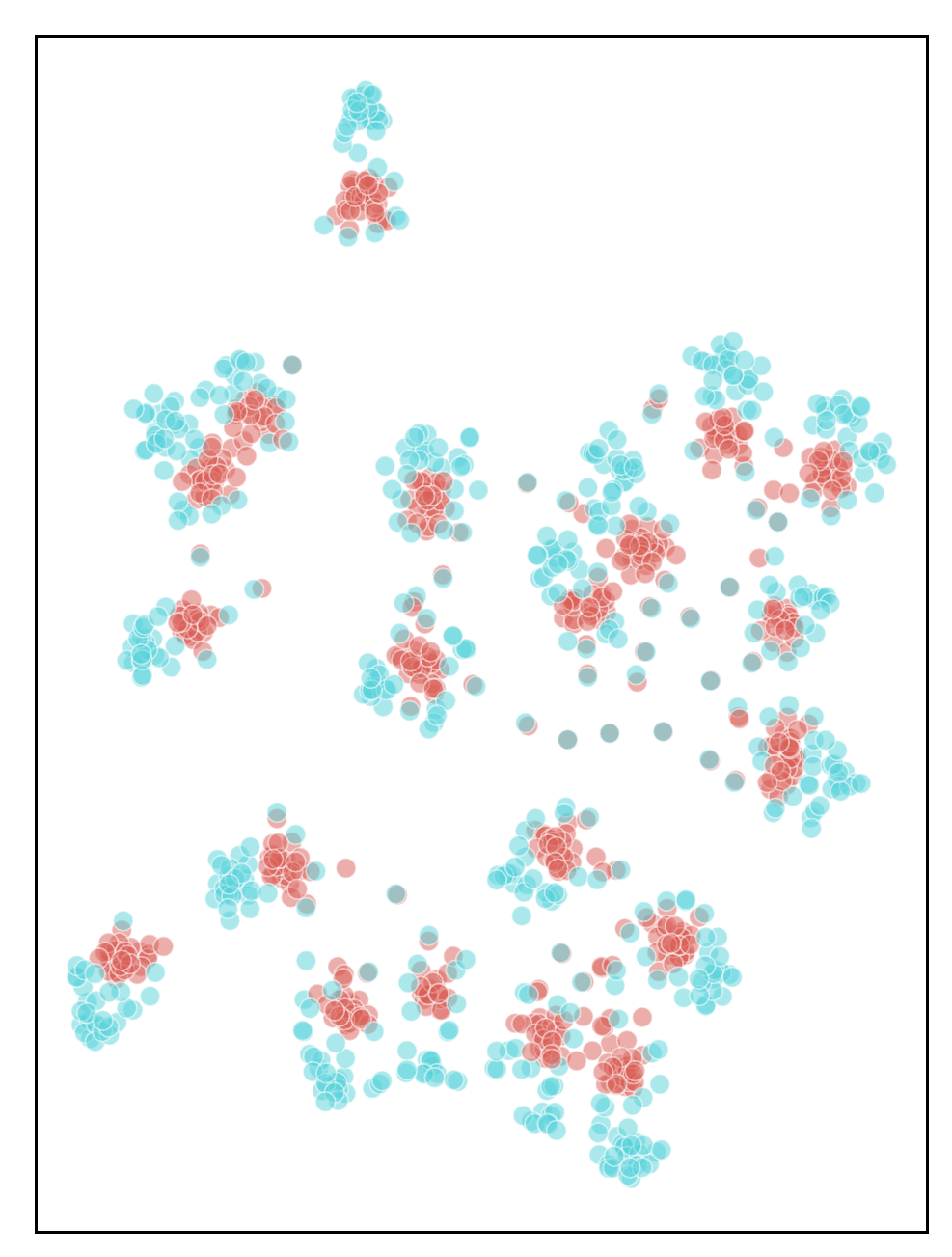}\\
        \fontsize{5cm}{5cm}\selectfont{(a) Baseline} &\fontsize{5cm}{5cm}\selectfont{(b) CS-KD} &\fontsize{5cm}{5cm}\selectfont{(c) TF-KD} &\fontsize{5cm}{5cm}\selectfont{(d) PS-KD} &\fontsize{5cm}{5cm}\selectfont{(e) AI-KD}\\
    \end{tabular}}
    \caption{t-SNE by arbitrarily selected 20 classes on ResNet-18 with CUB200-2011. In the top row, different colors indicate each class. In the bottom row, red and blue dots indicate the extracted features from the student and the superior model, respectively.}
    \label{fig:tsne}
\end{figure*}

We visualize the benefits of our proposed method using t-SNE~\cite{vandermaaten08a}. Figure~\ref{fig:tsne} shows the t-SNE visualization examples compared to representative methods on CUB200-2011 dataset with arbitrarily selected 20 classes for better visualization. 
The top row describes how well-clustered samples are in each class by the methods, and colors indicate different classes. AI-KD tends to have more extensive inter-class variances than other methods, showing outstanding classification performance.
The bottom row shows the dependence degree of the student model on the pre-trained model.
Blue and red dots indicate features extracted from $S^{Sup}$ and $S_{t}$, respectively. Because CS-KD does not utilize $S^{Sup}$, the pre-trained dots depict far from the student dots. 
In contrast, the t-SNE image by TF-KD shows that most red dots are overlapped with blue dots so that the student model is directly aligned to the pre-trained model and highly dependent on the pre-trained model.
Meanwhile, the visualized result by AI-KD shows that the student features (red dots) were placed near the pre-trained features (blue dots) or overlapped with blue dots partially. Also, AI-KD tends to concentrate on specific points of each class rather than blue dots. It represents AI-KD aligns the distribution of the student model to the one of the pre-trained model.
%-------------------------------------------------------------------------

%-------------------------------------------------------------------------
\subsection{Ablation studies on AI-KD}
\label{sec:exp_ablation}
In this section, we analyze the effectiveness of our proposed method using various ablation studies and experiments. First, we explore the impact of our proposed losses using an ablation study and then confirm the necessity of the adversarial loss function. By evaluating variants of the discriminator model, a suitable discriminator architecture is designated to lead to the best classification performance. Also, we experiment to search fitted temperature scaling parameters to better distill knowledge from $S^{Sup}$ and $S_{t-1}$ to $S_{t}$. Lastly, AI-KD has four proposed losses, and we conduct evaluations to find the balancing weights of each loss.

%-------------------------------------------------------------------------
\paragraph{Efficacy of the proposed losses}
AI-KD utilizes two loss functions with different objectives, which are $\mathcal{L}_{D}$ and $\mathcal{L}_{AI}$, as in Equations~\ref{eq:disc} and \ref{eq:tot}. $\mathcal{L}_{D}$ penalizes the discriminator to distinguish better the predictive distributions that are the pre-trained and the student models. $\mathcal{L}_{AI}$ supports regularizing $S_{t}$ using distilled information from $S^{Sup}$ and $S_{t-1}$.
To verify the benefits of the proposed adversarial and progressive learning scheme, we derive four variants of the proposed method, named ‘No Adv’, ‘Only Progressive’, ‘Only Guide’, and ‘Baseline’. ‘No Adv’ is to train the student model with the pre-trained model without adversarial learning, ‘Only Progressive’ denotes that the student model is trained by the progressive regularization $\mathcal{L}_{P}$ without pre-trained model, ‘Only Guide’ is implemented with the pre-trained model and the guide loss $\mathcal{L}_{G}$, and $\mathcal{L}_{G}$, and ‘Baseline’ refers to the basic student model trained only with the standard cross-entropy, $\mathcal{L}_{CE}$ for $S_t$. \cref{tab:abl_loss_weight_a} shows the effectiveness of the proposed method compared to its four variants on CIFAR-100 dataset with ResNet-18 in terms of Top-1 error and Top-5 error. The baseline recorded the worst performance and the proposed AI-KD outperformed the four variants in terms of Top-1 and Top-5 errors. As shown in Table~\ref{tab:abl_loss_weight_a} on the third and fourth row, we confirm the effectiveness of the progressive regularization and guide loss by the improved performances from the baseline by $+1.80\%$ and $+2.71\%$ in Top-1 error rates, respectively. In addition, ‘No Adv’ provided the second-best result in terms of Top-1 and Top-5 errors. Through employing the adversarial learning scheme, $S_t$ not only can learn to mimic the predictive probabilities of $S^{Sup}$ but also align the distributions between $S^{Sup}$ and $S_{t}$. In this manner, AI-KD leads to outperforming the other variants and shows improved performance.
%-------------------------------------------------------------------------
\begin{table*}[t]
\begin{center}
\footnotesize
\setlength{\tabcolsep}{1.0pt}
\begin{tabular}{p{0.15\linewidth}p{0.3\linewidth}p{0.2\linewidth}P{0.15\linewidth}P{0.15\linewidth}}
    \toprule
    Model &Dataset &Method &Top-1 Err\fontsize{0.22cm}{0.22cm}\selectfont~($\%$) &Top-5 Err\fontsize{0.22cm}{0.22cm}\selectfont~($\%$)\\
    \midrule
    \midrule
    \multirow{5}{*}{ResNet-18} &\multirow{5}{*}{CIFAR-100} &AI-KD &\textbf{19.87\fontsize{0.22cm}{0.22cm}\selectfont$\pm$0.07} &\textbf{4.81\fontsize{0.22cm}{0.22cm}\selectfont$\pm$0.04}\\
    & &No Adv &\underline{20.34\fontsize{0.22cm}{0.22cm}\selectfont$\pm$0.28} &\underline{4.86\fontsize{0.22cm}{0.22cm}\selectfont$\pm$0.10}\\
    & &Only Progressive &21.26\fontsize{0.22cm}{0.22cm}\selectfont$\pm$0.06 &5.66\fontsize{0.22cm}{0.22cm}\selectfont$\pm$0.06\\  
    & &Only Guide &22.17\fontsize{0.22cm}{0.22cm}\selectfont$\pm$0.05 &6.21\fontsize{0.22cm}{0.22cm}\selectfont$\pm$0.22\\
    & &Baseline &23.97\fontsize{0.22cm}{0.22cm}\selectfont$\pm$0.29 &6.80\fontsize{0.22cm}{0.22cm}\selectfont$\pm$0.02\\
    \bottomrule
\end{tabular}
\end{center}
\caption{Loss ablation study result with ResNet-18 on CIFAR-100. We report the mean and standard deviation over three runs. The best and second-best results are indicated in bold and underlined. Each loss which is $\mathcal{L}_{CE}$, $\mathcal{L}_{G}$, $\mathcal{L}_{P}$, and $\mathcal{L}_{A}$, are indicated the loss of the cross-entropy, guide, previous epoch, and adversarial, respectively.}
\label{tab:abl_loss_weight_a}
\end{table*}
%-------------------------------------------------------------------------
\begin{table}[!t]
\begin{center}
\footnotesize
\setlength{\tabcolsep}{1.0pt}
\begin{tabular}{p{0.15\linewidth}p{0.2\linewidth}P{0.3\linewidth}P{0.15\linewidth}P{0.15\linewidth}}
    \toprule
    Model &Dataset &$\omega$ &Top-1 Err\fontsize{0.22cm}{0.22cm}\selectfont~($\%$) &Top-5 Err\fontsize{0.22cm}{0.22cm}\selectfont~($\%$)\\
    \midrule
    \midrule
    \multirow{8}{*}{ResNet-18} &\multirow{8}{*}{CIFAR-100} &0.00 &20.34\fontsize{0.22cm}{0.22cm}\selectfont$\pm$0.28 &4.86\fontsize{0.22cm}{0.22cm}\selectfont$\pm$0.10\\
    & &0.01 &\underline{19.91\fontsize{0.22cm}{0.22cm}\selectfont$\pm$0.17} &\underline{4.82\fontsize{0.22cm}{0.22cm}\selectfont$\pm$0.14}\\
    & &0.05 &19.97\fontsize{0.22cm}{0.22cm}\selectfont$\pm$0.22 &4.85\fontsize{0.22cm}{0.22cm}\selectfont$\pm$0.25\\
    & &\textbf{0.10} &\textbf{19.87\fontsize{0.22cm}{0.22cm}\selectfont$\pm$0.07} &\textbf{4.81\fontsize{0.22cm}{0.22cm}\selectfont$\pm$0.04}\\
    & &0.25 &19.97\fontsize{0.22cm}{0.22cm}\selectfont$\pm$0.09 &4.89\fontsize{0.22cm}{0.22cm}\selectfont$\pm$0.09\\
    & &0.50 &20.17\fontsize{0.22cm}{0.22cm}\selectfont$\pm$0.12 &4.90\fontsize{0.22cm}{0.22cm}\selectfont$\pm$0.22\\
    & &0.75 &20.42\fontsize{0.22cm}{0.22cm}\selectfont$\pm$0.04 &5.27\fontsize{0.22cm}{0.22cm}\selectfont$\pm$0.29\\
    & &1.00 &20.42\fontsize{0.22cm}{0.22cm}\selectfont$\pm$0.04 &5.27\fontsize{0.22cm}{0.22cm}\selectfont$\pm$0.29\\
    \bottomrule
\end{tabular}
\end{center}
\caption{Ablation of adversarial loss weight ($\omega$) result on ResNet-18 with CIFAR-100. We report the mean and standard deviation over three runs. The best and second-best results are indicated in bold and underlined, respectively.}
\label{tab:abl_loss_weight_b}
\end{table}
%-------------------------------------------------------------------------
\begin{table}[!t]
\begin{center}
\footnotesize
\setlength{\tabcolsep}{1.0pt}
\begin{tabular}{p{0.15\linewidth}p{0.25\linewidth}P{0.09\linewidth}P{0.09\linewidth}P{0.09\linewidth}P{0.01\linewidth}P{0.15\linewidth}P{0.15\linewidth}}
    \toprule
    \multirow{2}{*}[-2pt]{~Model} &\multirow{2}{*}[-2pt]{~Dataset} &\multicolumn{3}{c}{Weight} &  &\multirow{2}{*}[-2pt]{Top-1 Err ($\%$)} &\multirow{2}{*}[-2pt]{Top-5 Err ($\%$)}\\
    \cmidrule{3-5}
    & &$\alpha_{CE}$ &$\alpha_{G}$ &$\alpha_{P}$ & & &\\
    \midrule
    \midrule
    \multirow{6}{*}{~ResNet-18} &\multirow{6}{*}{~CIFAR-100} &0.6 &0.1 &0.3 &  &\textbf{19.87\fontsize{0.22cm}{0.22cm}\selectfont$\pm$0.07} &\textbf{4.81\fontsize{0.22cm}{0.22cm}\selectfont$\pm$0.04}\\
    & &0.6 &0.2 &0.2 & &21.29\fontsize{0.22cm}{0.22cm}\selectfont$\pm$0.12 &5.46\fontsize{0.22cm}{0.22cm}\selectfont$\pm$0.09\\
    & &0.6 &0.3 &0.1 & &22.05\fontsize{0.22cm}{0.22cm}\selectfont$\pm$0.17 &5.62\fontsize{0.22cm}{0.22cm}\selectfont$\pm$0.11\\
    \cmidrule{3-8}
    & &0.7 &0.1 &0.2 & &20.98\fontsize{0.22cm}{0.22cm}\selectfont$\pm$0.09 &5.24\fontsize{0.22cm}{0.22cm}\selectfont$\pm$0.12\\
    & &0.7 &0.2 &0.1 & &21.79\fontsize{0.22cm}{0.22cm}\selectfont$\pm$0.13 &5.48\fontsize{0.22cm}{0.22cm}\selectfont$\pm$0.07\\
    \cmidrule{3-8}
    & &0.8 &0.1 &0.1 & &21.46\fontsize{0.22cm}{0.22cm}\selectfont$\pm$0.11 &5.16\fontsize{0.22cm}{0.22cm}\selectfont$\pm$0.20\\
    \bottomrule
\end{tabular}
\end{center}
\caption{Balanced weight ablation results based with ResNet-18 on CIFAR-100. We report the mean and standard deviation over three runs. The best results are indicated in bold.}
\label{tab:abl_weight_c}
\end{table}

%-------------------------------------------------------------------------
\paragraph{Balancing weight parameter of the losses}
We experiment with the adversarial loss variants, which are $\omega \in \{0.0, 0.01, 0.1, 0.5, 1.0\}$ to set proper weight. As shown in Table~\ref{tab:abl_loss_weight_b}, the classification performance is not sensitive to $\omega$ ranging from 0.01 to 0.5, though $\omega=0.1$ yields the best performance in the experiment.
Also, the loss of AI-KD consists of four losses that are cross-entropy loss, guide loss, progressive loss, and adversarial loss, denoted $\mathcal{L}_{CE}$, $\mathcal{L}_{G}$, $\mathcal{L}_{P}$, and $\mathcal{L}_{D}$, respectively. And the losses have balancing weights, such as $\alpha_{CE}$, $\alpha_{G}$, $\alpha_{P}$, and $\omega$. As we confirmed in the main paper, the weight of adversarial loss, $\omega$ is optimized to 0.1. In this section, we confirm other weights, $\alpha_{CE}$, $\alpha_{G}$, $\alpha_{P}$. Since these parameters are balancing weight, the sum of weight should be 1. We verify the variants as described in Table~\ref{tab:abl_weight_c}, AI-KD shows the best performance where the weight of $\alpha_{CE}$, $\alpha_{G}$, and $\alpha_{P}$ are 0.6, 0.1, and 0.3, respectively.

%-------------------------------------------------------------------------
\subsection{Compatibility with data augmentations}
\label{sec:exp_data_augmentations}
Self-KD is one of the regularization methods. And the regularization can be combined with other regularization techniques orthogonally. For this reason, we conducted compatibility experiments with well-known data augmentation techniques, such as Cutout~\cite{devries2017improved}, Mixup~\cite{zhang2018mixup}, and CutMix~\cite{yun2019cutmix}.
Cutout removes a randomly selected square region in images. Mixup generates additional data from existing training data by combining samples with an arbitrary mixing rate. CutMix combines Mixup and Cutout methods and generates new samples by replacing patches of random regions cropped from other images. As shown in Table~\ref{tab:aug}, we observe that Top-1 error rate records 19.75$\%$, 19.29$\%$, and 18.97$\%$ when our method is combined with Cutout, Mixup, and CutMix, respectively.
Additionally, all results present improved performance than AI-KD without data augmentation techniques.
It shows the broad applicability of our method for use with other regularization methods.
%-------------------------------------------------------------------------
\begin{table}[!t]
\begin{center}
\footnotesize
\setlength{\tabcolsep}{1.0pt}
\begin{tabular}{p{0.15\linewidth}p{0.25\linewidth}p{0.3\linewidth}P{0.15\linewidth}P{0.15\linewidth}}
    \toprule
    ~Model &~Dataset &~~Method &Top-1 Err ($\%$) &Top-5 Err ($\%$)\\
    \midrule
    \midrule
    \multirow{4}{*}{~ResNet-18} &\multirow{4}{*}{~CIFAR-100} &~~AI-KD &19.87\fontsize{0.22cm}{0.22cm}\selectfont$\pm$0.07 &4.81\fontsize{0.22cm}{0.22cm}\selectfont$\pm$0.04\\
    & &~~~+ Cutout~\cite{devries2017improved} &19.75\fontsize{0.22cm}{0.22cm}\selectfont$\pm$0.29 &4.90\fontsize{0.22cm}{0.22cm}\selectfont$\pm$0.01\\
    & &~~~+ Mixup~\cite{zhang2018mixup} &19.29\fontsize{0.22cm}{0.22cm}\selectfont$\pm$0.16 &4.70\fontsize{0.22cm}{0.22cm}\selectfont$\pm$0.12\\
    & &~~~+ CutMix~\cite{yun2019cutmix} &18.97\fontsize{0.22cm}{0.22cm}\selectfont$\pm$0.25 &4.32\fontsize{0.22cm}{0.22cm}\selectfont$\pm$0.10\\
    \bottomrule
\end{tabular}
\end{center}
\caption{Compatibility results with data augmentation based on ResNet-18 with CIFAR-100. We report the mean and standard deviation over three runs.}
\label{tab:aug}
\end{table}

%-------------------------------------------------------------------------
\section{Conclusion}
\label{sec:conclusion}
In this paper, we presented a novel self-knowledge distillation method using adversarial learning to improve the generalization of DNNs. 
The proposed method, AI-KD, distills deterministic and progressive knowledge from superior model's and the previous student's so that the student model learns stable. 
By applying adversarial learning, the student learns not only to mimic the corresponding network but also to consider aligning the distributions from the superior pre-trained model.
Compared to representative Self-KD methods, AI-KD presented better image classification performances in terms of Top-1 error, Top-5 Error, and F1 score with network architectures on various datasets.
In particular, our proposed method recorded outstanding performances on fine-grained datasets.
In addition, the ablation study showed the advantages of the proposed adversarial learning and implicit regularization.
Also, we demonstrated the compatibility with data augmentation methods.
Regardless, AI-KD still has a limitation which is requiring the pre-trained model. The pre-trained model is essential since it mimics distributions and utilizes adversarial learning. However, the requirement hinders end-to-end learning.

For future work, we look forward to proposing a KD method without the pre-trained model, so that developing sustainable DNNs. Furthermore, we plan to extend AI-KD to continual learning and multimodal machine learning.

%-------------------------------------------------------------------------
\section*{Acknowledgement}
This work was funded by Samsung Electro-Mechanics and was partially supported by Carl-Zeiss Stiftung under the Sustainable Embedded AI project (P2021-02-009).

%-------------------------------------------------------------------------
%%APA style
% \bibliographystyle{model5-names}\biboptions{authoryear}
%\bibliographystyle{apalike}
\bibliography{article}

%-------------------------------------------------------------------------
% \newpage
% \appendix
% \section{Visualizing Experimental Result with Bar Graphs}
% \label{visualization}

%-------------------------------------------------------------------------
\end{document}